\documentclass[onecolumn,11pt]{article}
\usepackage[top=1in, bottom=1in, left=1in, right=1in]{geometry}
\usepackage{amsfonts,amssymb,amsmath}

\usepackage[numbers,sort&compress]{natbib}
\usepackage[bottom,flushmargin,hang,multiple]{footmisc}

\usepackage{graphicx}        

\usepackage[boxed]{algorithm2e}
\usepackage{mathtools}
\usepackage{xcolor}
\usepackage{url,hyperref,lineno,microtype}
\usepackage{appendix}

\mathtoolsset{showonlyrefs}		
\graphicspath{{figures/}}

\newcommand{\R}{\mathbb{R}}

\newcommand{\x}{\mathbf{x}}
\newcommand{\X}{\mathbf{X}}
\newcommand{\f}{\mathbf{f}}
\newcommand{\bXi}{\mathbf{\Xi}}
\newcommand{\bPhi}{\mathbf{\Phi}}
\newcommand{\bbm}{\begin{bmatrix}}
\newcommand{\ebm}{\end{bmatrix}}
\newcommand{\paren}[1]{\left(#1\right)}
\newcommand{\norm}[1]{\left\|#1\right\|}

\DeclareMathOperator*{\argmin}{arg\,min}

\makeindex             

\newcommand\blfootnote[1]{%
  \begingroup
  \renewcommand\thefootnote{}\footnote{#1}%
  \addtocounter{footnote}{-1}%
  \endgroup
}

\usepackage{listings}

\definecolor{maroon}{cmyk}{0, 0.87, 0.68, 0.32}
\definecolor{halfgray}{gray}{0.55}
\definecolor{ipython_frame}{RGB}{207, 207, 207}
\definecolor{ipython_bg}{RGB}{247, 247, 247}
\definecolor{ipython_red}{RGB}{186, 33, 33}
\definecolor{ipython_green}{RGB}{0, 128, 0}
\definecolor{ipython_cyan}{RGB}{64, 128, 128}
\definecolor{ipython_purple}{RGB}{170, 34, 255}

\lstdefinelanguage{iPython}{
    morekeywords={access,and,break,class,continue,def,del,elif,else,except,exec,finally,for,from,global,if,import,in,is,lambda,not,or,pass,print,raise,return,try,while,True,False,as},%
    morekeywords=[2]{abs,all,any,basestring,bin,bool,bytearray,callable,chr,classmethod,cmp,compile,complex,delattr,dict,dir,divmod,enumerate,eval,execfile,file,filter,float,format,frozenset,getattr,globals,hasattr,hash,help,hex,id,input,int,isinstance,issubclass,iter,len,list,locals,long,map,max,memoryview,min,next,object,oct,open,ord,pow,property,range,raw_input,reduce,reload,repr,reversed,round,set,setattr,slice,sorted,staticmethod,str,sum,super,tuple,type,unichr,unicode,vars,xrange,zip,apply,buffer,coerce,intern},%
    sensitive=true,%
    morecomment=[l]\#,%
    morestring=[b]',%
    morestring=[b]",%
    morestring=[s]{'''}{'''},
    morestring=[s]{"""}{"""},
    morestring=[s]{r'}{'},
    morestring=[s]{r"}{"},%
    morestring=[s]{r'''}{'''},%
    morestring=[s]{r"""}{"""},%
    morestring=[s]{u'}{'},
    morestring=[s]{u"}{"},%
    morestring=[s]{u'''}{'''},%
    morestring=[s]{u"""}{"""},%
    literate=
    {á}{{\'a}}1 {é}{{\'e}}1 {í}{{\'i}}1 {ó}{{\'o}}1 {ú}{{\'u}}1
    {Á}{{\'A}}1 {É}{{\'E}}1 {Í}{{\'I}}1 {Ó}{{\'O}}1 {Ú}{{\'U}}1
    {à}{{\`a}}1 {è}{{\`e}}1 {ì}{{\`i}}1 {ò}{{\`o}}1 {ù}{{\`u}}1
    {À}{{\`A}}1 {È}{{\'E}}1 {Ì}{{\`I}}1 {Ò}{{\`O}}1 {Ù}{{\`U}}1
    {ä}{{\"a}}1 {ë}{{\"e}}1 {ï}{{\"i}}1 {ö}{{\"o}}1 {ü}{{\"u}}1
    {Ä}{{\"A}}1 {Ë}{{\"E}}1 {Ï}{{\"I}}1 {Ö}{{\"O}}1 {Ü}{{\"U}}1
    {â}{{\^a}}1 {ê}{{\^e}}1 {î}{{\^i}}1 {ô}{{\^o}}1 {û}{{\^u}}1
    {Â}{{\^A}}1 {Ê}{{\^E}}1 {Î}{{\^I}}1 {Ô}{{\^O}}1 {Û}{{\^U}}1
    {œ}{{\oe}}1 {Œ}{{\OE}}1 {æ}{{\ae}}1 {Æ}{{\AE}}1 {ß}{{\ss}}1
    {ç}{{\c c}}1 {Ç}{{\c C}}1 {ø}{{\o}}1 {å}{{\r a}}1 {Å}{{\r A}}1
    {€}{{\EUR}}1 {£}{{\pounds}}1,
    literate=
    *{-}{{{\color{ipython_purple}-}}}1
     {?}{{{\color{ipython_purple}?}}}1
     {^}{{{\color{ipython_purple}\^{}}}}1
     {=}{{{\color{ipython_purple}=}}}1
     {+}{{{\color{ipython_purple}+}}}1
     {*}{{{\color{ipython_purple}$^\ast$}}}1
     {/}{{{\color{ipython_purple}/}}}1
     {>}{{{\color{ipython_purple}$>$}}}1
     {<}{{{\color{ipython_purple}$<$}}}1
     {@}{{{\color{ipython_purple}@}}}1
     {+=}{{{+=}}}1
     {-=}{{{-=}}}1
     {*=}{{{$^\ast$=}}}1
     {/=}{{{/=}}}1,
    identifierstyle=\color{black}\ttfamily,
    commentstyle=\color{ipython_cyan}\ttfamily,
    stringstyle=\color{ipython_red}\ttfamily,
    keepspaces=true,
    showspaces=false,
    showstringspaces=false,
    rulecolor=\color{ipython_frame},
    frame=single,
    frameround={t}{t}{t}{t},
    framexleftmargin=6mm,
    numbers=left,
    numberstyle=\tiny\color{halfgray},
    backgroundcolor=\color{ipython_bg},
    basicstyle=\scriptsize,
    keywordstyle=\color{ipython_green}\ttfamily,
}

\title{\vspace{-.55in}{\fontsize{16}{16}\selectfont \textbf{Discovery of Physics from Data:  Universal Laws and Discrepancies}}\vspace{-.15in}}

\author{\normalsize{Brian M. de Silva$^{1*}$, David M. Higdon$^2$, Steven L. Brunton$^3$, J. Nathan Kutz$^1$}\\
\footnotesize{$^1$ Department of Applied Mathematics, University of Washington, Seattle, WA}\\
\footnotesize{$^2$ Department of Statistics, Viginia Polytechnic Institute and State University, Blacksburg, VA}\\
\footnotesize{$^3$ Department of Mechanical Engineering, University of Washington, Seattle, WA\vspace{-.2in}}
}

\date{}

\begin{document}
\maketitle

\blfootnote{$^*$ Corresponding author (bdesilva@uw.edu).}

\vspace{-0.2in}
\begin{abstract}
	Machine learning (ML) and artificial intelligence (AI) algorithms are now being used to automate the discovery of physics principles and governing equations from measurement data alone.  
	However, positing a universal physical law from data is challenging without simultaneously proposing an accompanying discrepancy model to account for the inevitable mismatch between theory and measurements. 
	By revisiting the classic problem of modeling falling objects of different size and mass, we highlight a number of nuanced issues that must be addressed by modern data-driven methods for automated physics discovery.
	Specifically, we show that measurement noise and complex secondary physical mechanisms, like unsteady fluid drag forces, can obscure the underlying law of gravitation, leading to an erroneous model.  
	We use the sparse identification of nonlinear dynamics (SINDy) method to identify governing equations for real-world measurement data and simulated trajectories.
	Incorporating into SINDy the assumption that each falling object is governed by a similar physical law is shown to improve the robustness of the learned models, but discrepancies between the predictions and observations persist due to subtleties in drag dynamics. 
	This work highlights the fact that the naive application of ML/AI will generally be insufficient to infer universal physical laws without further modification.%

	\vspace{0.15in}
\end{abstract}

\section{Introduction}
	The ability to derive governing equations and physical principles has been a hallmark feature of scientific discovery and technological progress throughout human history.
	Even before the scientific revolution, the Ptolemaic doctrine of the \textit{perfect circle}~\cite{ptolemy,peters1915ptolemy} provided a principled decomposition of planetary motion into a hierarchy of circles, i.e. a bona fide theory for planetary motion.
	The scientific revolution and the resulting development of calculus provided the mathematical framework and language to precisely describe scientific principles, including gravitation, fluid dynamics, electromagnetism, quantum mechanics, etc.
	With advances in data science over the past few decades, principled methods are emerging for such scientific discovery from time-series measurements alone.
	Indeed, across the engineering, physical and biological sciences, significant advances in sensor and measurement technologies have afforded unprecedented new opportunities for scientific exploration.
	Despite its rapid advancements and wide-spread deployment, \textit{machine learning} (ML) and \textit{artificial intelligence} (AI) algorithms for scientific discovery face significant challenges and limitations, including noisy and corrupt data, latent variables, multiscale physics, and the tendency for overfitting.
	In this manuscript, we revisit one of the classic problems of physics considered by Galileo and Newton, that of falling objects and gravitation.
	We demonstrate that a sparse regression framework is well-suited for physics discovery, while highlighting both the need for principled methods to extract parsimonious physics models and the challenges associated with the naive application of ML/AI techniques.
	Even this simplest of physical examples demonstrates critical principles that must be considered in order to make data-driven discovery viable across the sciences.

	Measurements have long provided the basis for the discovery of governing equations.
	Through empirical observations of planetary motion, the Ptolemaic theory of motion was developed~\cite{ptolemy,peters1915ptolemy}.
	This was followed by Kepler's laws of planetary motion and the elliptical courses of planets in a heliocentric coordinate system~\cite{kepler2015astronomia}.
	By hand calculation, he was able to regress Brahe's state-of-the-art data on planetary motion to the minimally parametrized elliptical orbits which described planetary orbits with a terseness the Ptolemaic system had never managed to achieve.
	Such models led to the development of Newton's ${\bf F}=m{\bf a}$~\cite{newton1999principia}, which provided a universal, generalizable, interpretable, and succinct description of physical dynamics.
	Parsimonious models are critical in the philosophy of Occam's razor: the simplest set of explanatory variables is often the best~\cite{blumer1987occam,domingos1999role,Bongard2007pnas,Schmidt2009science}.
	It is through such models that many technological and scientific advancements have been made or envisioned.

	What is largely unacknowledged in the scientific discovery process is the intuitive leap required to formulate physics principles and governing equations.
	Consider the example of falling objects.
	According to physics folklore, Galileo discovered, through experimentation, that objects fall with the same constant acceleration, thus disproving Aristotle's theory of gravity, which stated that objects fall at different speeds depending on their mass.
	The leaning tower of Pisa is often the setting for this famous stunt, although there is little evidence such an experiment actually took place~\cite{cooper1936aristotle,adler1978galileo,segre1980role}.
	Indeed, many historians consider it to have been a thought experiment rather than an actual physical test.
	Many of us have been to the top of the leaning tower and have longed to drop a bowling ball from the top, perhaps along with a golf ball and soccer ball, in order to replicate this experiment.
	If we were to perform such a test, here is what we would likely find: Aristotle was correct.
	Balls of different masses and sizes \textit{do} reach the ground at different times.
	As we will show from our own data on falling objects, (noisy) experimental measurements may be insufficient for discovering a constant gravitational acceleration, especially when the objects experience Reynolds numbers varying by orders of magnitudes over the course of their trajectories.
	But what is beyond dispute is that Galileo did indeed \textit{posit} the idea of a fixed acceleration, a conclusion that would have been exceptionally difficult to come to from such measurement data alone.
	Gravitation is only one example of the intuitive leap required for a paradigm shifting physics discovery.
	Maxwell's equations~\cite{maxwell1873treatise} have a similar story arc revolving around Coulomb's inverse square law.
	Maxwell cited Coulomb's torsion balance experiment as establishing the inverse square law while dismissing it only a few pages later as an approximation~\cite{falconer2017no,bartlett1970experimental}.
	Maxwell concluded that Faraday's observation that an electrified body, touched to the inside of a conducting vessel, transfers all its electricity to the outside surface as much more direct proof of the square law.
	In the end, both would have been approximations, with Maxwell taking the intuitive leap that exactly a power of negative two was needed when formulating Maxwell's equations.
	Such examples abound across the sciences, where intuitive leaps are made and seminal theories result.

	One challenge facing ML and AI methods is their inability to take such leaps.
	At their core, many ML and AI algorithms involve regressions based on data, and are statistical in nature~\cite{breiman2001statistical,wu2008top,bishop2006pattern,murphy2012machine}.
	Thus by construction, a model based on measurement data would not produce an exact inverse {\it square} law, but rather a slightly different estimate of the exponent.
	In the case of falling objects, ML and AI would yield an Aristotelian theory of gravitation, whereby the data would suggest that objects fall at a speed related to their mass.
	Of course, even Galileo intuitively understood that air resistance plays a significant role in the physics of falling objects, which is likely the reason he conducted controlled experiments on inclined ramps.
	Although we understand that air resistance, which is governed by latent fluid dynamic variables, explains the discrepancy between the data and a constant gravity model, our algorithms do not.
	Without modeling these small disparities (e.g., due to friction, heat dissipation, air resistance, etc.), it is almost impossible to uncover universal laws such as gravitation.
	Differences between theory and data have played a foundational role in physics, with general relativity arising from inconsistencies between gravitational theory and observations, and quantum mechanics arising from our inability to explain the photoelectric effect with Maxwell's equations.

	Our goal in this manuscript is to highlight the many subtle and nuanced concerns related to data-driven discovery using modern ML and AI methods.
	Specifically, we highlight these issues on the most elementary of problems: modeling the motion of falling objects.
	Given our ground-truth knowledge of the physics, this example provides a convenient testbed for different physics discovery techniques.
	It is important that one clearly understands the potential pitfalls in such methods before applying them to more sophisticated problems which may arise in fields like biology, neuroscience, and climate modeling.
	Our physics discovery method is rooted in the \textit{sparse identification for nonlinear dynamics} (SINDy) algorithm, which has been shown to extract parsimonious governing equations in a broad range of physical sciences~\cite{Brunton2016pnas}.
	SINDy has been widely applied to identify models for fluid flows~\cite{Loiseau2017jfm,Loiseau2018jfm}, optical systems~\cite{Sorokina2016oe}, chemical reaction dynamics~\cite{Hoffmann2018arxiv},  convection in a plasma~\cite{Dam2017pf}, structural modeling~\cite{lai2019sparse}, and for model predictive control~\cite{Kaiser2018prsa}.
	There are also a number of theoretical extensions to the SINDy framework, including for identifying partial differential equations~\cite{Rudy2017sciadv,Schaeffer2017prsa},  and models with rational function nonlinearities~\cite{Mangan2016ieee}.
	It can also incorporate partially known physics and constraints~\cite{Loiseau2017jfm}.
	The algorithm can be reformulated to include integral terms for noisy data~\cite{Schaeffer2017pre} or handle incomplete or limited data~\cite{Tran2016arxiv,schaeffer2018extracting}.
	In this manuscript we show that \emph{group sparsity}~\cite{Rudy2018arxiv} may be used to enforce that the same model terms explain \emph{all} of the observed trajectories, which is essential in identifying the correct model terms without overfitting.

	SINDy is by no means the only attempt that has been made at using machine learning to infer physical models from data.
	Gaussian processes have been employed to learn conservation laws described by parametric linear equations~\cite{raissi2017jcp}.
	Symbolic regression has been successfully applied to the problem of inferring dynamics from data~\cite{Bongard2007pnas,Schmidt2009science}.
	Another closely related set of approaches are process-based models \cite{tanevski2017ProcessBasedModeling,bridewell2008ml,tanevski2016bmc} which, similarly to SINDy, allow one to specify a library of relationships or functions between variables based on domain knowledge and produce an interpretable set of governing equations.
	The principal difference between process-based models and SINDy is that SINDy employs sparse regression techniques to perform function selection which allows a larger class of library functions to be considered than is tractable for process-based models.
	Deep learning methods have been proposed for accomplishing a variety of related tasks such as predicting physical dynamics directly~\cite{mrowca2018nips}, building neural networks that respect given physical laws~\cite{raissi2017arxivPt1}, discovering parameters in nonlinear partial differential equations with limited measurement data~\cite{raissi2017arxivPt2}, and simultaneously approximating the solution and nonlinear dynamics of nonlinear partial differential equations ~\cite{raissi2018jmlr}.
	Graph neural networks~\cite{battaglia2018arxiv}, a specialized class of neural networks that operate on graphs, have been shown to be effective at learning basic physics simulators from measurement data~\cite{battaglia2016nips,chang2016iclr} and directly from videos~\cite{watters2017nips}.
	It should be noted that the aforementioned neural network approaches either require detailed prior knowledge of the form of the underlying differential equations or fail to yield simple sets of interpretable governing equations.

\section{Materials and Methods}

	\subsection{Fluid forces on a sphere: A brief history}
		It must have been immediately clear to Galileo and Newton that committing to a gravitational constant created an inconsistency with experimental data.
		Specifically, one had to explain why objects of different sizes and shapes fall at different speeds (e.g. a feather versus a cannon ball).
		Wind resistance was an immediate candidate to explain the \textit{discrepancy} between a universal gravitational constant and measurement data.
		The fact that Galileo performed experiments where he rolled balls down inclines seems to suggest that he was keenly aware of the need to isolate and disambiguate the effects of gravitational forces from fluid drag forces.
		Discrepancies between the Newtonian theory of gravitation and observational data of Mercury's orbit led to Einstein's development of general relativity.
		Similarly, the photoelectric effect was a discrepancy in Maxwell's equations which led to the development of quantum mechanics.

\begin{figure}
			\centering
			\includegraphics[width=.65\textwidth]{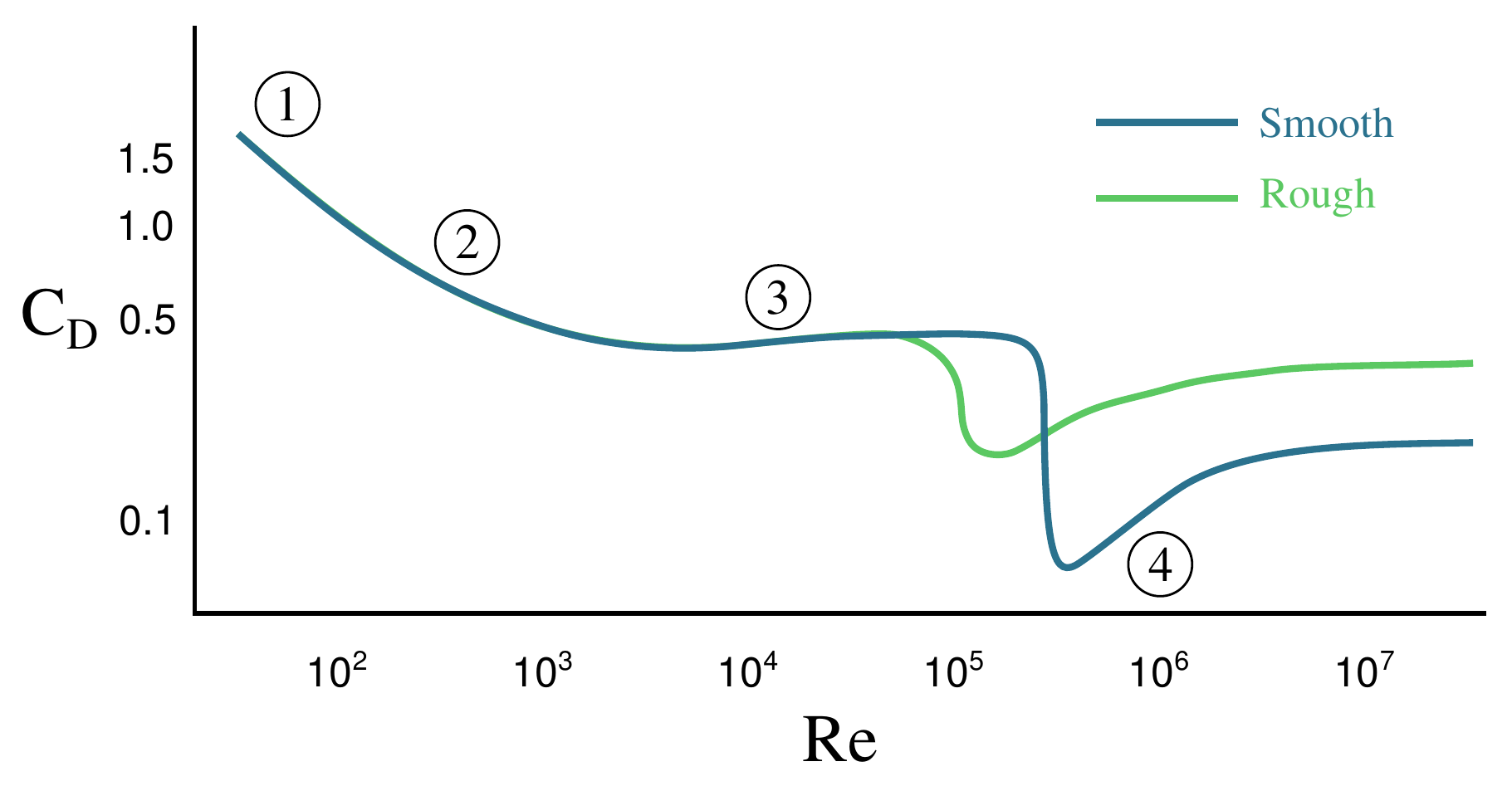}
			\caption{The drag coefficient for a sphere as a function of Reynolds number, $Re$. The dark curve shows the coefficient for a sphere with a smooth surface and the light curve a sphere with a rough surface. The numbers highlight different flow regimes. (1) attached flow and steady separated flow; (2) separated unsteady flow, with laminar flow boundary layer upstream of separation, producing a K\'arm\'an vortex street; (3) separated unsteady flow with a chaotic turbulent wake downstream and a laminar boundary layer upstream; (4) post-critical separated flow with turbulent boundary layer.}
			\label{fig:SphereDrag}
		\end{figure}

		Discrepancy modeling is therefore a critical aspect of building and discovering physical models.
		Consider the motion of falling spheres as a prototypical example.
		In addition to the force of gravity, a falling sphere encounters a fluid drag force as it passes through the air.
		A simple model of the drag force $F_D$ is given by:
		\begin{align}\label{Eq:fluiddrag}
		F_D = \frac{1}{2}\rho v^2 A C_D,
		\end{align}
		where $\rho$ is the fluid density, $v$ is the velocity of the sphere with respect to the fluid, $A=\pi D^2 / 4$ is the cross-sectional area of the sphere, $D$ is the diameter of the sphere, and $C_D$ is the dimensionless drag coefficient.  
		As the sphere accelerates through the fluid, its velocity increases, exciting various unsteady aerodynamic effects, such as laminar boundary layer separation, vortex shedding, and eventually a turbulent boundary layer and wake~\cite{moller1938experimentelle,magarvey1965vortices,calvert1972some,achenbach1972experiments,achenbach1974vortex,smits2004aerodynamics}.  
		Thus, the drag coefficient is a function of the sphere's velocity, and this coefficient generally decreases for increasing velocity.
		Figure~\ref{fig:SphereDrag} shows the drag coefficient $C_D$ for a sphere as a function of the Reynolds number $Re=\rho v D/\mu$, where $\mu$ is the dynamic viscosity of the fluid; for a constant diameter and viscosity, the Reynolds number is directly proportional to the velocity.  
		Note that the drag coefficient of a smooth sphere will differ from that of a rough sphere.  
		The flow over a rough sphere will become turbulent at lower velocities, causing less flow separation and a more streamlined, lower-drag wake; this explains why golf balls are dimpled, so that they will travel farther~\cite{smits2004aerodynamics}. 
		Thus, \eqref{Eq:fluiddrag} states that drag is related to the square of the velocity, although $C_D$ has a weak dependence on velocity.
		When $Re$ is small, $C_D$ is proportional to $1/v$, resulting in a drag force that is linear in $v$. For larger values of $Re$, $C_D$ is approximately constant (away from the steep drop), leading to a quadratic drag force.
		Eventually, the drag force will balance the force of gravity, resulting in the sphere reaching its \emph{terminal velocity}. 
		In addition, as the fluid wake becomes unsteady, the drag force will also vary in time, although these variations are typically fast and may be time-averaged.   
		Finally, objects accelerating in a fluid will also accelerate the fluid out of the way, resulting in an effective mass that includes the mass of the body and an \emph{added mass} of accelerated fluid~\cite{Newman:1977}; however, this added mass force will typically be quite small in air.  

		In addition to the theoretical study of fluid forces on an idealized sphere, there is a rich history of scientific inquiry into the aerodynamics of sports balls \cite{smits2004aerodynamics,Goff2013SportsEngineering,mehta2008sports,mehta1985FluidMechanicsReview}.
		Apart from gravity and drag, a ball’s trajectory can be influenced by the spin of the ball via the Magnus force or lift force which acts in a direction orthogonal to the drag.
		Other factors that can affect the forces experienced by a falling ball include air temperature, wind, elevation, and ball surface shape.

	\subsection{Data set}\label{sec:dataset}

\begin{figure}[!ht]
			\centering
			\includegraphics[width=\textwidth]{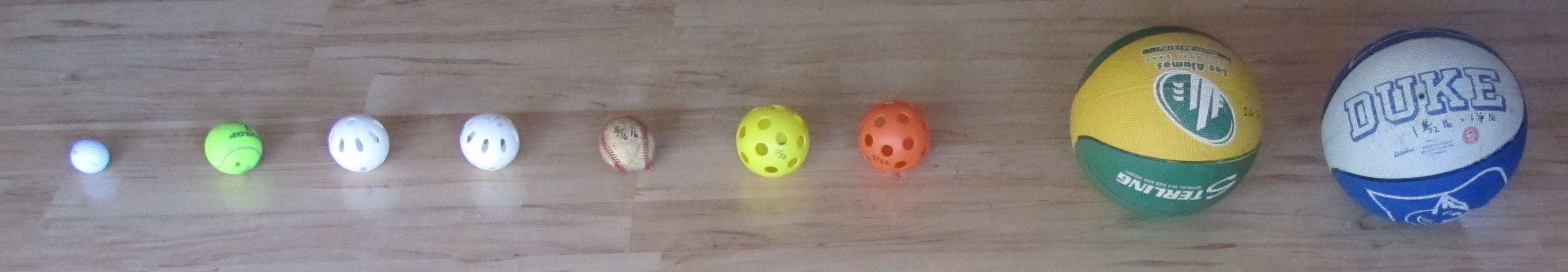}
			\caption{The balls that were dropped from the bridge, with the volleyball omitted. From left to right: Golf Ball, Tennis Ball, Whiffle Ball 1, Whiffle Ball 2, Baseball, Yellow Whiffle Ball, Orange Whiffle Ball, Green Basketball, and Blue Basketball. The two colored whiffle balls have circular openings and are structurally identical. The two white whiffle balls have elongated slits and are also identical.}
			\label{fig:balls}
		\end{figure}

\begin{figure}[!ht]
			\centering
			\includegraphics[width=\textwidth]{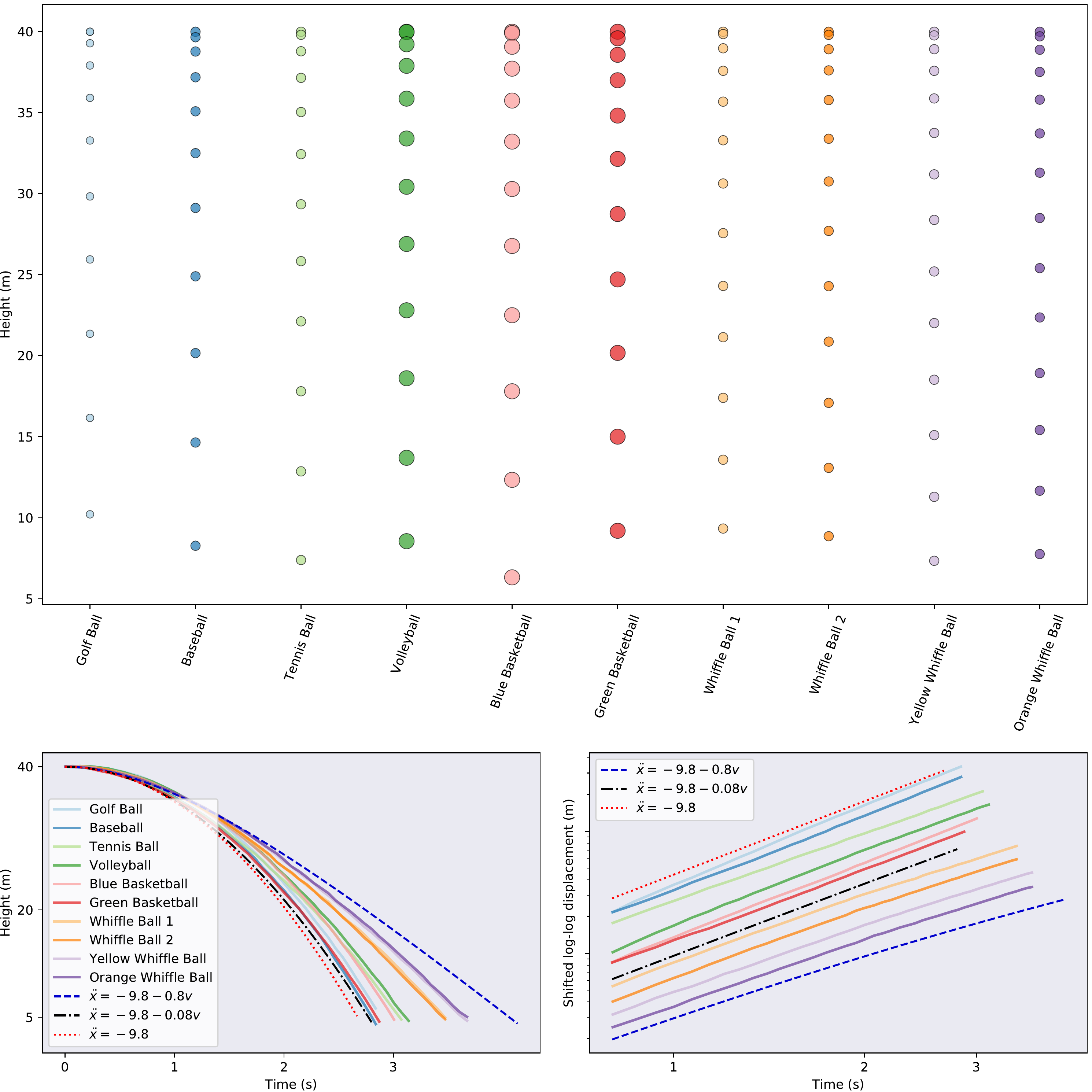}
			\caption{Visualizations of the ball trajectories for the second drop. Top: Subsampled raw drop data for each ball. Bottom left: Height for each ball as a function of time. We also include the simulated trajectories of idealized balls with differing levels of drag (black and blue) and a ball with constant acceleration (red). Bottom right: A log-log plot of the displacement of each ball from its original position atop the bridge. Note that we have shifted the curves vertically and zoomed in on the later segments of the time series to enable easier comparison. In this plot a ball falling at a constant rate (zero acceleration) will have a trajectory represented by a line with slope one. A ball falling with constant acceleration will have a trajectory represented by a line with slope two. A ball with drag will have a trajectory which begins with slope two and asymptotically approaches a line with slope one.}
			\label{fig:ball_drops}
		\end{figure}

		The data considered in this manuscript are height measurements of balls falling through air.
		These measurements originate from two sources: physical experiments and simulations.
		Such experiments are popular in undergraduate physics classes where they are used to explore linear versus quadratic drag \cite{owen2005ejp,kaewsutthi2011student,cross2014physicsteacher,christensen2014physicseducation} and scaling laws \cite{sznitman2017ejpe}.
		In June 2013 a collection of balls, pictured in Figure \ref{fig:balls}, were dropped, {\it twice each}, from the Alex Fraser Bridge in Vancouver, BC from a height of about 35 meters above the landing site.
		In total 11 balls were dropped: a golf ball, a baseball, two whiffle balls with elongated holes, two whiffle balls with circular holes, two basketballs, a bowling ball, and a volleyball (not pictured).
		More information about the balls is given in Table \ref{tab:ball-stats}.
		The air temperature at the time of the drops was 65 degrees Fahrenheit (18$^{\circ}$ Celsius).
		A hand held iPad was used to record video of the drops at a rate of 15 frames per second.
		The height of the falling objects was then estimated by tracking the balls in the resulting videos.
		Figure \ref{fig:ball_drops} visualizes the second set of ball drops.
		As one might expect, the whiffle balls all reach the ground later than the other balls.
		This is to be expected since the openings in their faces increase the drag they experience.
		Even so, all the balls reach the ground within a second of each other.
		We also plot the simulated trajectories of two spheres falling with constant linear (in $v$) drag and the trajectory predicted by constant acceleration.
		Note that, based on the log-log plot of displacement, none of the balls appears to have reached terminal velocity by the time they hit the ground.
		This may increase the difficulty of accurately inferring the balls' governing equations.
		Given only measurements from one regime of falling ball dynamics, it may prove difficult to infer models that generalize to other regimes.

\begin{figure}[!ht]
			\centering
			\includegraphics[width=\textwidth]{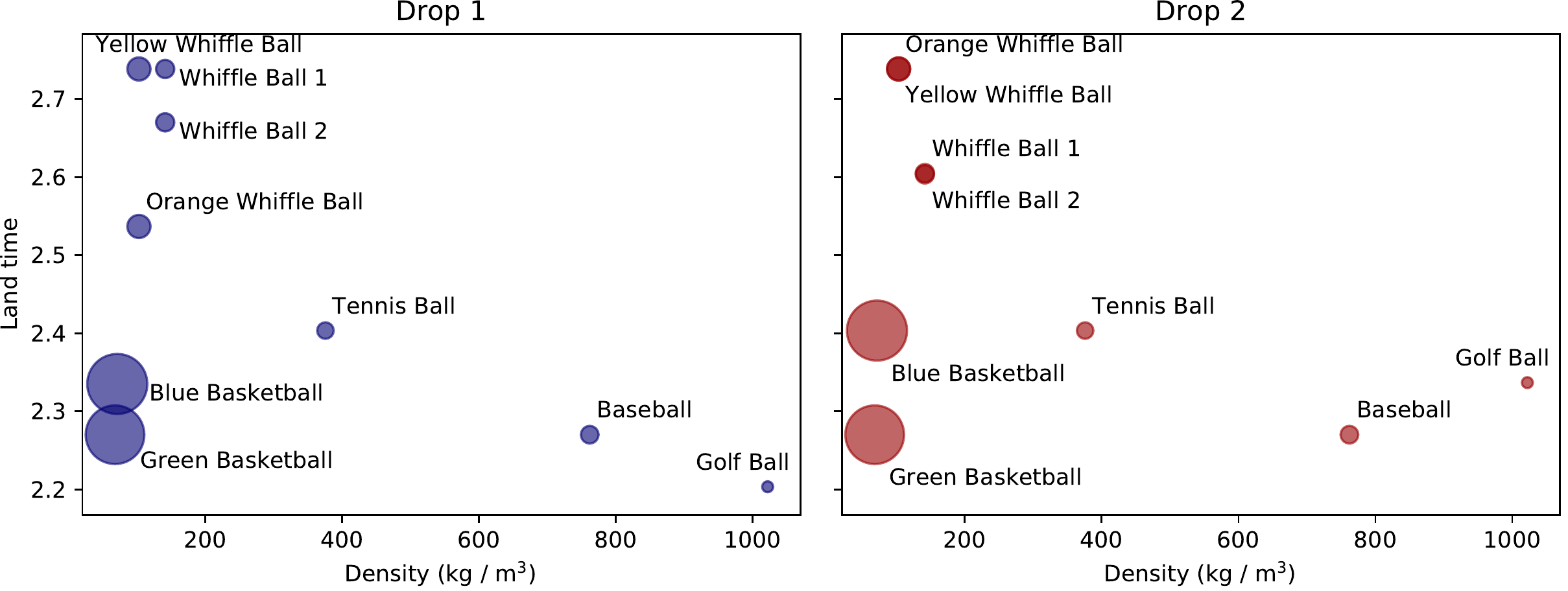}
			\caption{The amount of time taken by each ball to travel a fixed distance as a function of ball density.}
			\label{fig:landing_time_vs_density}
		\end{figure}

		Drawing inspiration from Aristotle, one might form the hypothesis that the amount of time taken by spheres to reach the ground should be a function of the \textit{density} of the spheres.
		Density takes into account both information about the mass of an object and its volume, which might be thought to affect the air resistance it encounters.
		We plot the landing time of each ball as a function of its density for both drops in Figure \ref{fig:landing_time_vs_density}.
		To be more precise, because some balls were dropped from slightly different heights, we measure the amount of time it takes each ball to travel a fixed distance after being dropped, not the amount of time it takes the ball to reach the ground.
		There is a general trend across the tests for the denser balls to travel faster.
		However, the basketballs defy this trend and complete their journeys about as quickly as the densest ball.
		This shows there must be more factors at play than just density.
		There is also variability in the land time of the balls across drops.
		While most of the balls have very consistent fall times across drops, the blue basketball, golf ball, and orange whiffle ball reach the finish line faster in the first trial than the second one.
		These differences could be due to a variety of factors, including the balls being released with different initial velocities, or errors in measuring the balls' heights.


		There are multiple known sources of error in the measurement data.
		The relatively low resolution of the videos means that the inferred ball heights are only approximate.
		In the appendix (Section \ref{sec:estimating-noise}) we attempt to infer the level of noise introduced by our use of heights derived from imperfect video data.
		Furthermore, the camera was held by a person, not mounted on a tripod, leading to shaky footage.
		The true bridge height is uncertain because it was measured with a laser range finder claiming to be accurate to within 0.5 meters.
		Because the experiments were executed outside, it is possible for any given drop to have been affected by wind.
		Detecting exactly when each ball was dropped, at what velocity it was dropped, and when it hit the ground using only videos is certain to introduce further error.
		Finally, treating these balls as perfect spheres is an approximation whose accuracy depends on the nature of the balls.
		This idealization seems least appropriate for the whiffle balls, which are sure to exhibit much more complicated aerodynamic effects than, say, the baseball.
		The bowling ball was excluded from consideration because of corrupted measurements from its first drop.

		The situation we strive to mimic with this experiment is one in which the researcher is in a position of ignorance about the system being studied.
		In order to design an experiment which eliminates the effects of confounding factors such as air resistance one must already have an appreciation for which factors are worth controlling; one leverages prior knowledge as Galileo did when he employed ramps in his study of falling objects to mitigate the effect of air resistance.
		In the early stages of investigation of a physical phenomenon, one must often perform poorly-controlled experiments to help identify these factors.
		We view the ball drop trials as this type of experiment.

		In addition to the measurement data just described, we construct a synthetic data set by simulating falling objects with masses of 1 kg and different (linear) drag coefficients.
		In particular, for each digital ball, we simulate two drops of the same length as the real data and collect height measurements at a rate of 15 measurements per second.
		The balls fall according to the equation $\ddot x(t) = -9.8 + D\dot x(t)$, with each ball having its own \textit{constant} drag coefficient, $D<0$.
		We simulate five balls in total, with respective drag coefficients $-0.1$, $-0.3$, $-0.3$, $-0.5$, and $-0.7$.
		These coefficients are all within the plausible range suggested by the simulated trajectories shown in Figure \ref{fig:ball_drops}.
		Each object is ``dropped'' with an initial velocity of 0.
		Varying amounts of Gaussian noise are added to the height data so that we may better explore the noise tolerance of the proposed model discovery approaches:
		\begin{equation}
			\tilde x_i = x_i + \eta \epsilon_i.
		\end{equation}
		where $\eta \geq 0$ and $\epsilon_i\sim N(0,1)$; that is to say $\epsilon_i$ is normally distributed with unit variance.

	\subsection{Methods}
		
		In this section we describe the model discovery methods we employ to infer governing equations from noisy data.
		We first give the mathematical background necessary for learning dynamics via sparse regression and provide a brief overview of the SINDy method in Section \ref{sec:SINDy}.
		In Section \ref{sec:group-sparsity} we propose a group sparsity regularization strategy for improving the robustness and generalizability of SINDy.
		We briefly discuss the setup of the model discovery problem we are attempting to solve in Section \ref{sec:eqs-of-motion}.
		Finally, we discuss numerical differentiation, a subroutine critical to effective model discovery, in Section \ref{sec:numerical-differentiation}.
		\subsubsection{Sparse identification of nonlinear dynamical systems}\label{sec:SINDy}
			Consider the nonlinear dynamical system for the state vector $\x(t)=[x_1(t),x_2(t),\dots,x_n(t)]^\top\in\R^n$ defined by
			\begin{equation}
				\dot\x = \f(\x(t)).
			\end{equation}
			Given a set of noisy measurements of $\x(t)$, the sparse identification of nonlinear dynamics (SINDy) method, introduced in~\cite{Brunton2016pnas}, seeks to identify $\f:\R^n\to\R^n$. In this section we give an overview of the steps involved in the SINDy method and the assumptions upon which it relies. Throughout this manuscript we refer to this algorithm as the \textit{unregularized SINDy} method, not because it involves no regularization, but because its regularization is not as closely tailored to the problem at hand as the method proposed in Section \ref{sec:group-sparsity}. 

			For many dynamical systems of interest, the function specifying the dynamics, $\f$, consists of only a few terms. That is to say, when represented in the appropriate basis, there is a sense in which it is sparse. The key idea behind the SINDy method is that if one supplies a rich enough set of candidate functions for representing $\f$, then the correct terms can be identified using sparse regression techniques. The explicit steps are as follows. First we collect a set of (possibly noisy) measurements of the state $\x(t)$ and its derivative $\dot \x(t)$ at a sequence of points in time, $t_1, t_2,\dots, t_m$. These measurements are concatenated into two matrices, the columns of which correspond to different state variables and the rows of which correspond to points in time.
			\begin{align}
				\X &= \bbm \x(t_1)^\top \\ \x(t_2)^\top \\ \vdots \\ \x(t_m)^\top \ebm
				 = \bbm
				 		x_1(t_1) & x_2(t_1) & \dots & x_n(t_1) \\
				 		x_1(t_2) & x_2(t_2) & \dots & x_n(t_2) \\
				 		\vdots & \vdots & \ddots & \vdots \\
				 		x_1(t_m) & x_2(t_m) & \dots & x_n(t_m)
				 	\ebm, \\
				\dot \X &=\bbm \dot\x(t_1)^\top \\ \dot\x(t_2)^\top \\ \vdots \\ \dot\x(t_m)^\top \ebm
				 = \bbm
				 		\dot x_1(t_1) & \dot x_2(t_1) & \dots & \dot x_n(t_1) \\
				 		\dot x_1(t_2) & \dot x_2(t_2) & \dots & \dot x_n(t_2) \\
				 		\vdots & \vdots & \ddots & \vdots \\
				 		\dot x_1(t_m) & \dot x_2(t_m) & \dots & \dot x_n(t_m)
				 	\ebm.
			\end{align}

			Next we specify a set of candidate functions, $\{\phi_i(\x): i=1,2,\dots,p\}$, with which to represent $\f$. Examples of candidate functions include monomials up to some finite degree, trigonometric functions, and rational functions. In practice the selection of these functions can be informed by the practitioner's prior knowledge about the system being measured. The candidate functions are evaluated on $\X$ to construct a library matrix
			\begin{equation}
				\bPhi(\X) = \bbm \vrule & \vrule & & \vrule \\
						\phi_1(\X) & \phi_2(\X) & \dots & \phi_p(\X) \\
						\vrule & \vrule & & \vrule \ebm.
			\end{equation}
			Note that each column of $\bPhi(\X)$ corresponds to a single candidate function. Here we have overloaded notation and interpret $\phi(\X)$ as the column vector obtained by applying $\phi_i$ to each row of $\X$. It is assumed that each component of $\f$ can be represented as a \textit{sparse} linear combination of such functions. This allows us to pose a regression problem to be solved for the coefficients used in these linear combinations:
			\begin{equation}\label{eq:full-sindy-eqs}
				\dot \X = \bPhi(\X)\bXi.
			\end{equation}
			We adopt MATLAB-style notation and use $\bXi_{(:,j)}$ to denote the $j$-th column of $\bXi$. The coefficients specifying the dynamical system obeyed by $\x_j$ are stored in $\bXi_{(:,j)}$:
			\begin{equation}
				\dot \x_j = \f_j(\x) = \bPhi\paren{\x^\top}\bXi_{(:,j)},
			\end{equation}
			where $\bPhi\paren{\x^\top}$ is to be interpreted as a (row) vector of symbolic functions of components of $\x$. The full system of differential equations is then given by
			\begin{equation}
				\dot \x = \f(\x) = \bXi^\top \paren{\bPhi\paren{\x^\top}}^\top.
			\end{equation}

			For concreteness we supply the following example. With the candidate functions $\left\{1, \allowbreak x_1, \allowbreak x_2, \allowbreak x_1x_2, \allowbreak x_1^2, \allowbreak x_2^2\right\}$ the Lotka-Volterra equations
			\begin{equation}
				\begin{cases}
					\dot x_1 =& \alpha x_1 - \beta x_1 x_2, \\
					\dot x_2 =& \delta x_1 x_2 - \gamma x_2
				\end{cases}
			\end{equation}
			can be expressed as 
			\begin{equation}
				\dot \x =
				\bbm \dot x_1 \\ \dot x_2 \ebm =
				\bXi^\top \paren{\bPhi\paren{\x^\top}}^\top = 
				\bbm
					0 & \alpha & 0 & -\beta & 0 & 0\\ 0 & 0 & -\gamma & \delta & 0 & 0
				\ebm
				\bbm
					1 \\ x_1 \\ x_2 \\ x_1 x_2 \\ x_1^2 \\ x_2^2
				\ebm
			\end{equation}

			Were we to obtain pristine samples of $\x(t)$ and $\dot \x(t)$ we could solve \eqref{eq:full-sindy-eqs} exactly for $\bXi$. Furthermore, assuming we chose linearly independent candidate functions and avoided collecting redundant measurements, $\bXi$ would be unique and would exhibit the correct sparsity pattern. In practice, however, measurements are contaminated by noise and we actually observe a perturbed version of $\x(t)$.
			In many cases $\dot \x(t)$ is not observed directly and must instead be approximated from $\x(t)$, establishing another source of error. The previously exact equation, \eqref{eq:full-sindy-eqs}, to be solved for $\bXi$ is supplanted by the approximation problem
			\begin{equation}
				\dot \X \approx \bPhi(\X)\bXi.
			\end{equation}
			To find $\bXi$ we solve the more concrete optimization problem
			\begin{equation}\label{eq:obj-fcn}
				\min_{\bXi}\frac12 \norm{\dot \X - \bPhi(\X)\bXi}_F^2 + \Omega(\bXi),
			\end{equation}
			where $\Omega(\cdot)$ is a regularization term chosen to promote sparse solutions and $\|\cdot\|_F$ is the Frobenius norm.
			Note that because any given column of $\bXi$ encodes a differential equation for a single component of $\x$, each column generates a problem that is decoupled from the problems associated with the other columns.
			Thus, solving \eqref{eq:obj-fcn} consists of solving $n$ separate regularized least squares problems.
			Row $i$ of $\bXi$ contains the coefficients of library function $\phi_i$ for each governing equation.

			The most direct way to enforce sparsity is to choose $\Omega$ to be the $\ell_0$ penalty, defined as $\|\mathbf{M}\|_0 = \sum_{i,j}|\text{sign}(M_{ij})|$.
			This penalty simply counts the number of nonzero entries in a matrix or vector.
			However, using the $\ell_0$ penalty makes \eqref{eq:obj-fcn} difficult to optimize because $\norm{\cdot}_0$ is nonsmooth and nonconvex.
			Another common choice is the $\ell_1$ penalty defined by $\|\mathbf{M}\|_1 = \sum_{i,j}|M_{ij}|$.
			This function is the convex relaxation of the $\ell_0$ penalty.
			The LASSO, proposed in~\cite{Tibshirani1996rss}, with coordinate descent is typically employed to solve $\eqref{eq:obj-fcn}$ with $\Omega(\cdot)=\|\cdot\|_1$, but this method can become computationally expensive for large data sets and often leads to incorrect sparsity patterns~\cite{su2017false}.
			Hence we solve \eqref{eq:obj-fcn} using the sequential thresholded least-squares algorithm proposed in~\cite{Brunton2016pnas}, and studied in further detail in~\cite{zheng2018unified}.
			In essence, the algorithm alternates between (a) successively solving the \textit{unregularized} least-squares problem for each column of $\bXi$ and (b) removing candidate functions from consideration whose corresponding components in $\bXi$ are below some threshold.
			This threshold or sparsity parameter, is straightforward to interpret: no governing equations are allowed to have any terms with coefficients of magnitude smaller than the threshold.
			Crucially, it should be noted that just because a candidate function is discarded for one column of $\bXi$ (i.e. for one component's governing equation) does not mean it is removed from contention for the other columns.
			A simple Python implementation of sequentially thresholded least-squares is provided in the appendices (Section \ref{sec:sindy-algo})
			
			We note that if we simulate falling objects with constant acceleration, $\ddot x(t) =  -9.8$, or linear drag, $\ddot x(t) = -9.8 + D\dot x(t)$, and add \textit{no noise}, then there is almost perfect agreement between the true governing equations and the models learned by SINDy.
			The appendices contains a more thorough discussion of such numerical experiments and another example application of SINDy.

			SINDy has a number of well-known limitations.
			The biggest of these is the effect of noise on the learned equations.
			If one does not have direct measurements of derivatives of state variables, then these derivatives must be computed numerically.
			Any noise that is present in the measurement data is amplified when it is numerically differentiated, leading to noise in both $\dot \X$ and $\bPhi(\X)$ in \eqref{eq:obj-fcn}.
			In its original formulation, SINDy often exhibits erratic performance in the face of such noise, but extensions have been developed which handle noise more gracefully \cite{Schaeffer2017pre,Tran2016arxiv}.
			We discuss numerical differentiation further in Section \ref{sec:numerical-differentiation}.
			As with other methods, each degree of freedom supplied to the practitioner presents a potential source of difficulty.
			To use SINDy one must select a set of candidate functions, a sparse regularization function, and a parameter weighing the relative importance of the sparseness of the solution against accuracy.
			An improper choice of any one of these can lead to poor performance.
			The set of possible candidate functions is infinite, but SINDy requires one to specify a finite number of them. If one has any prior knowledge of the dynamics of the system being modeled, it can be leveraged here.
			If not, it is typically recommended to choose a class of functions general enough to encapsulate a wide variety of behaviors (e.g. polynomials or trigonometric functions).
			In theory, sparse regression techniques should allow one to specify a sizable library of functions, selecting only the relevant ones.
			However, in practice, the underlying regression problem becomes increasingly ill-conditioned as more functions are added.
			If one wishes to explore an especially large space of possible library functions it may be better to use other approaches such as symbolic regression with genetic algorithms \cite{Bongard2007pnas,Schmidt2009science}.
			A full discussion of how to pick a sparsity-promoting regularizer is beyond the scope of this work.
			We do note that there have been recent efforts to explore different methods for obtaining sparse solutions when using SINDy \cite{champion2019unified}.
			An appropriate value for the sparsity hyperparameter can be obtained using cross-validation.
			We note that the need to perform hyperparameter tuning is by no means unique to SINDy.
			Virtually all machine learning methods require some amount of hyperparameter tuning. 
			There are two natural options for target metrics during cross-validation.
			The derivatives directly predicted by the linear model can be compared against the measured (or numerically computed) derivatives.
			Alternatively, the model can be fed into a numerical integrator along with initial conditions to obtain predicted future values for the state variables.
			These forecasts can then be judged against the measured values.
			To achieve a balance between model sparsity and accuracy, information theoretic criteria such as the Akaike information criteria (AIC) or Bayes information criteria (BIC) can be applied \cite{Mangan2017prsa}.

		\subsubsection{Group sparsity regularization}\label{sec:group-sparsity}
			The standard, unregularized SINDy approach attempts to learn the dynamics governing each state variable independently. It does not take into account prior information one may possess regarding relationships between state variables. Intuitively speaking, the balls in our data set (whiffle balls, perhaps, excluded) are similar enough objects that the equations governing their trajectories should include similar terms. In this subsection we propose a group sparsity method which can be interpreted as enforcing this hypothesis when seeking predictive models for the balls.

			We draw inspiration for our approach from the group LASSO of~\cite{Yuan2006rss}, which extends the LASSO. The classic LASSO method solves the $\ell_1$ regularization problem
			\begin{equation}\label{eq:lasso}
				\mathbf{\beta} = \argmin_{\mathbf{\beta}} \frac12 \norm{\X \mathbf{\beta} - \mathbf{Y}}_2^2 + \lambda \norm{\mathbf{\beta}}_1.
			\end{equation}
			which penalizes the magnitude of each component of $\beta$ \textit{individually}. The group LASSO approach modifies \eqref{eq:lasso} by bundling sets of related entries of $\mathbf{\beta}$ together when computing the penalty term. Let the entries of $\mathbf{\beta}$ be partitioned into $G$ disjoint blocks $\{\mathbf{\beta}_1, \mathbf{\beta}_2,\dots,\mathbf{\beta}_G\}$, which can be treated as vectors. The group LASSO then solves the following optimization problem
			\begin{equation}\label{eq:group-lasso}
				 	\mathbf{\beta} = \argmin_{\mathbf{\beta}} \frac12 \norm{\X \mathbf{\beta} - \mathbf{Y}}_2^2 + \lambda \sum_{i=1}^G \norm{\mathbf{\beta}_i}_2.
			\end{equation} 
			In the case that the groups each consist of exactly one entry of $\mathbf{\beta}$, \eqref{eq:group-lasso} reduces to \eqref{eq:lasso}.
			When blocks contain multiple entries, the group LASSO penalty encourages them to be retained or eliminated as a group.
			Furthermore, it drives sets of unimportant variables to truly vanish, unlike the $\ell_2$ regularization function which merely assigns small but nonzero values to insignificant variables.

			We apply similar ideas in our \textit{group sparsity} method for the SINDy framework and force the models learned for each ball to select the same library functions.
			Recall that the model variables are contained in $\bXi$.
			To enforce the condition that each governing equation should involve the same terms, we identify \textit{rows} of $\bXi$ as sets of variables to be grouped together.
			Borrowing MATLAB notation again, we let $\bXi_{(i,:)}$ denote row $i$ of $\bXi$.
			To perform sequential thresholded least squares with the group sparsity constraint we repeatedly apply the following steps until convergence: (a) solve the least-squares problem \eqref{eq:obj-fcn} \textit{without} a regularization term for each column of $\bXi$ (i.e. for each ball), (b) prune the library, $\bPhi(\X)$, of functions which have low relevance across most or all of the balls.
			This procedure is summarized in Algorithm \ref{alg:group-sparsity}.

			\begin{algorithm}[H]
				\KwData{$\dot \X\in\R^{m\times d}$, $\bPhi(\X)\in\R^{m\times p}$, and $\delta>0$}
				\KwResult{coefficient matrix $\bXi\in\R^{p\times d}$}
				\While{not converged}{
					\tcp{Solve a least squares problem for each state variable}
					\For{$j\gets 1$ \KwTo $d$}{
						$\bXi_{(:,j)} \gets \argmin_{\xi} \frac12 \norm{\dot\X - \bPhi(\X)\xi}_2^2$\;
					}
					\tcp{Remove library functions with low salience}
					\For{$i\gets 1$ \KwTo $p$}{
						\If{R$(\bXi_{(i,:)})<\delta$}{
							Delete $\bXi_{(i,:)}$ and $\bPhi(\X)_{(:,i)}$\;
						}
					}
				}
				Replace deleted rows of $\bXi$ and deleted columns of $\bPhi(\X)$ with 0's\;
				\caption{A group sparsity algorithm for the sequential thresholded least squares method}
				\label{alg:group-sparsity}
			\end{algorithm}

			Here $R$ is a function measuring the importance of a row of coefficients.
			Possible choices for $R$ include the $\ell_1$ or $\ell_2$ norm of the input, the mean or median of the absolute values of the entries of the input, or another statistical property of the input entries such as the lower 25\% quantile.
			In this work we use the $\ell_1$ norm.
			Convergence is attained when no rows of $\bXi$ are removed.
			Note that while all the models are constrained to be generated by the same library functions, the \textit{coefficients} in front of each can differ from one model to the next.
			The hyperparameter $\delta$ controls the sparsity of $\bXi$, though not as directly as the sparsity parameter for SINDy.
			Increasing it will result in models with fewer terms and decreasing it will have the opposite effect.
			Since we use the $\ell_1$ norm and there are 10 balls in our primary data set, rows of $\bXi$ whose average magnitude is less than $\tfrac{\delta}{10}$ are removed.

			Because the time series are all noisy, it is likely that some the differential equations returned by the unregularized SINDy algorithm will acquire spurious terms.
			Insisting that only terms which \textit{most} of the models find useful are kept, as with our group sparsity method, should help to mitigate this issue.
			In this way we are able to leverage the fact that we have multiple trials involving similar objects to improve the robustness of the learned models to noise.
			Even if some of the unregularized models from a given drop involve erroneous library functions, we might still hope that, on average, the models will pick the correct terms.
			Our approach can also be viewed as a type of \textit{ensemble} method wherein a set of models is formed from the time series of a given drop, they are allowed to vote on which terms are important, then the models are retrained using the constrained set of library functions agreed upon in the previous step.

		\subsubsection{Equations of motion}\label{sec:eqs-of-motion}
			Even the simplest model for the height, $x(t)$, of a falling object involves an acceleration term. Consequently, we impose the restriction that our model be a second order (autonomous) differential equation:
			\begin{equation}\label{eq:second-order}
				\ddot x = f(x, \dot x).
			\end{equation}
			The SINDy framework is designed to work with first order systems of differential equations, so we convert \eqref{eq:second-order} into such a system:
			\begin{equation}
				\begin{cases}
					\dot x =& v\\
					\dot v =& g(x,v).
				\end{cases}
			\end{equation}
			We then apply SINDy, with $\x=\bbm x & v\ebm^\top$ and $f(\x) = \bbm v & g(\x) \ebm^\top$, and attempt to learn the function $g$. In fact, because we already know the correct right-hand side function for $\dot x$, we need only concern ourselves with finding an expression for $\dot v$.

\begin{figure}
				\centering
				\includegraphics[width=\textwidth]{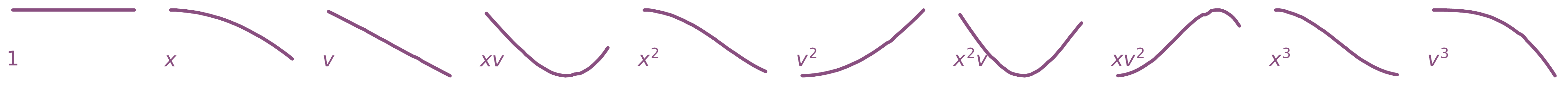}
				\caption{Visualizations of nonlinear library functions corresponding to the second green basketball drop. If the motion of the balls is described by Newton's second law, $F=m\ddot x$, then these functions can be interpreted as possible forcing terms constituting $F$.}
				\label{fig:library_row}
			\end{figure}

			Our nonlinear library consists of polynomials in $x$ and $v$ up to degree three, visualized in Figure \ref{fig:library_row}:
			\begin{equation}\label{eq:nonlinear-library}
				\bPhi(\X) = \bbm
				 \vrule & \vrule & \vrule & \vrule & \vrule & \vrule & & \vrule \\
				 \mathbf{1} & \mathbf{x}(t) & \mathbf{v}(t) & \mathbf{x}(t)\mathbf{v}(t) & \mathbf{x}(t)^2 & \mathbf{v}(t)^2 & \dots & \mathbf{v}(t)^3  \\ 
				 \vrule & \vrule & \vrule & \vrule & \vrule & \vrule & & \vrule
				\ebm.
			\end{equation}

			Assuming that the motion of the balls is completely determined by Newton's second law, $F=ma = m\ddot x$, we may interpret the SINDy algorithm as trying to discover the force (after dividing by mass) that explains the observed acceleration.

			Though we know now that the acceleration of a ball should not depend on its height, we seek to place ourselves in a position of ignorance analogous to the position scientists would have found themselves in centuries ago.
			We leave it to our algorithm to sort out which terms are appropriate.
			In practice one might selectively choose which functions to include in the library based on domain knowledge, or known properties of the system being modeled.

	 	\subsubsection{Numerical differentiation}\label{sec:numerical-differentiation}
	 	
		 	In order to form the nonlinear library \eqref{eq:nonlinear-library} and the derivative matrix, $\dot\X$, we must approximate the first two derivatives of the height data from each drop.
		 	Applying standard numerical differentiation techniques to a signal amplifies any noise that is present.
		 	This poses a serious problem since we aim to fit a model to the \textit{second} derivative of the height measurements.
		 	Because the amount of noise in our data set is nontrivial, two iterations of numerical differentiation will create an intolerable noise level.
		 	To mitigate this issue we apply a Savitzky-Golay filter from~\cite{Savtizky1964ac} to smooth the data before differentiating via second order centered finite differences.
		 	Points in a noisy data set are replaced by points lying on low-degree polynomials which are fit to localized patches of the original data with a least-squares method.
		 	Other available approaches include using a total variation regularized derivative as in~\cite{Brunton2016pnas} or working with an integral formulation of the governing equations as described in~\cite{Schaeffer2017pre}.
		 	We perform a detailed analysis of the error introduced by smoothing and numerical differentiation in an appendix (Section \ref{sec:numerical-differentiation-study}.

\section{Results}

	\subsection{Learned terms}
		In this section we compare the terms present in the governing equations identified using the unregularized SINDy approach with those present when the group sparsity constraint is imposed.
		We train separate models on the two drops.
		The two algorithms are given one sparsity hyperparameter each to be applied for all balls in both drops.
		The group sparsity method used a value of $1.5$ and the other method used a value of $0.04$.
		These parameters were chosen by hand to balance allowing the algorithms enough expressiveness to model the data, while being restrictive enough to prevent widespread overfitting;
		increasing them produces models with one or no terms and decreasing them results in models with large numbers of terms.
		See Appendix Section \ref{sec:sparsity-paramater-study} for a more detailed discussion of our choice of sparsity parameter values.

\begin{figure}
			\centering
			\includegraphics[width=\textwidth]{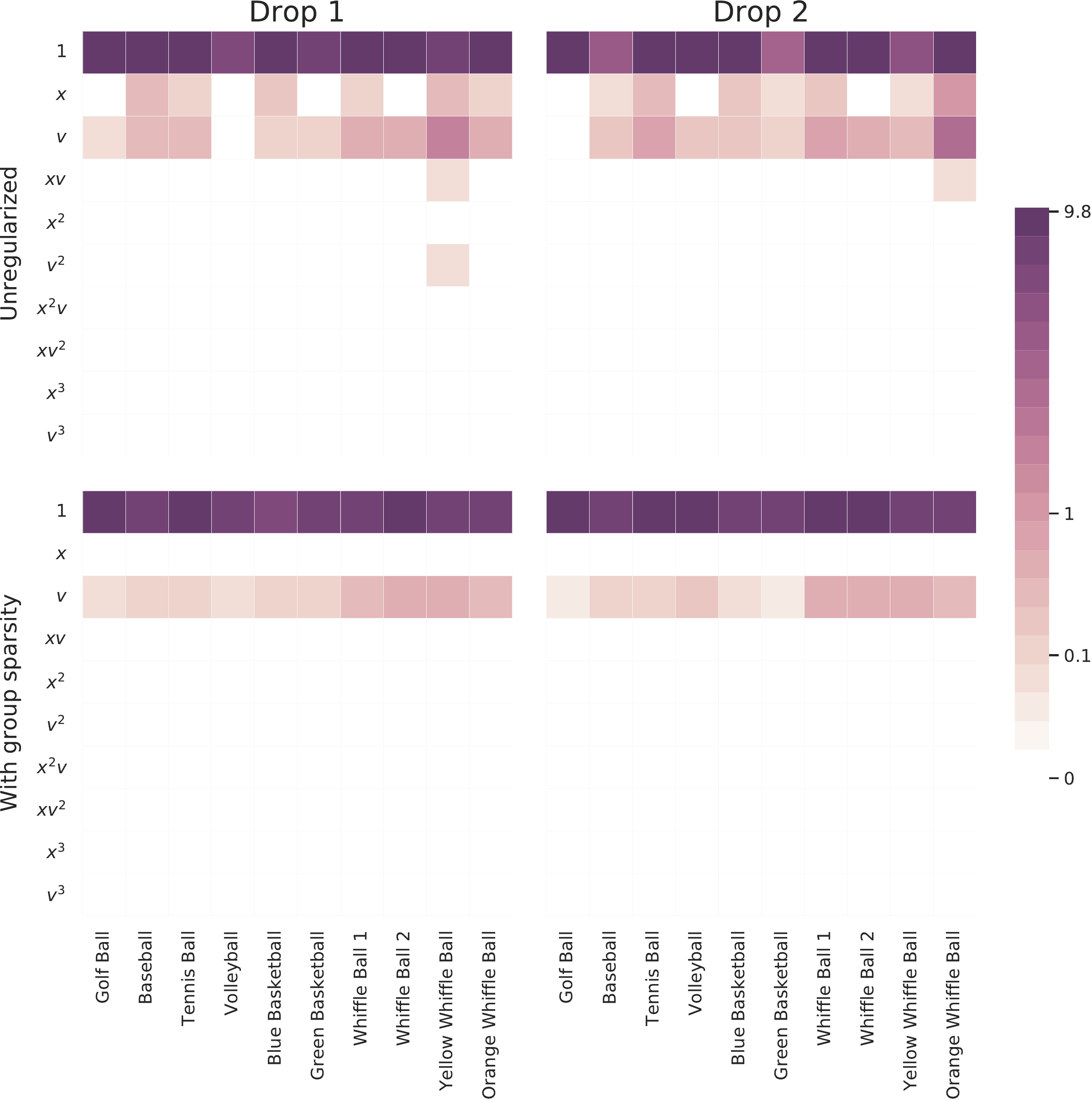}
			\caption{Magnitudes of the coefficients learned for each ball by models trained on one drop either with or without the proposed group sparsity approach. The unregularized approach used a sparsity parameter of $0.04$ and the group sparsity method used a value of $1.5$. Increasing this parameter slightly in the unregularized case serves to push many models to use only a constant function.}
			\label{fig:heatmaps}
		\end{figure}

		Figure \ref{fig:heatmaps} summarizes the results of this experiment.
		Learning a separate model for each ball independent of the others allows many models to fall prey to overfitting.
		Note how most of the governing equations incorporate an extraneous height term.
		On the other hand, two of the learned models involve only constant acceleration and fail to identify any effect resembling air resistance.

		The method leveraging group sparsity is more effective at eliminating extraneous terms and selecting only those which are useful across most balls.
		Moreover, only the constant and velocity terms are active, matching our intuition that the dominant forces at work are gravity and drag due to air resistance.
		Interestingly, the method prefers a linear drag term, one proportional to $v$, to model the discrepancy between measured trajectories and constant acceleration.
		Even the balls which don't include a velocity term in the unregularized model have this term when group sparsity regularization is employed.
		This shows that group penalty can simultaneously help to dismiss distracting candidate functions and promote correct terms that may have been overlooked.
		Is is also reassuring to see that, compared to the other balls, the whiffle ball models have larger coefficients on the $v$ terms.
		Their accelerations slow at a faster rate as a function of their velocities than do the other balls.

		\begin{table}[h]
			\centering
			\begin{tabular}{l|l|l}
				\textbf{Ball} & \textbf{First drop} & \textbf{Second drop}\\ \hline
				Golf Ball	        & $\ddot x=-9.34 +0.05v$	& $\ddot x=-9.44 -0.03v$	\\ 
				Baseball	        & $\ddot x=-8.51 +0.14v$	& $\ddot x=-7.56 +0.14v$	\\ 
				Tennis Ball	        & $\ddot x=-9.08 -0.13v$	& $\ddot x=-8.64 -0.12v$	\\ 
				Volleyball	        & $\ddot x=-8.11 -0.08v$	& $\ddot x=-9.64 -0.23v$	\\ 
				Blue Basketball	    & $\ddot x=-6.71 +0.15v$	& $\ddot x=-7.50 +0.07v$	\\ 
				Green Basketball	& $\ddot x=-7.36 +0.10v$	& $\ddot x=-8.05 +0.02v$	\\ 
				Whiffle Ball 1	    & $\ddot x=-8.24 -0.34v$	& $\ddot x=-9.44 -0.43v$	\\ 
				Whiffle Ball 2	    & $\ddot x=-9.81 -0.56v$	& $\ddot x=-9.79 -0.48v$	\\ 
				Yellow Whiffle Ball	& $\ddot x=-8.50 -0.47v$	& $\ddot x=-8.45 -0.46v$	\\ 
				Orange Whiffle Ball	& $\ddot x=-7.83 -0.35v$	& $\ddot x=-8.03 -0.42v$	\\ 
			\end{tabular}
			\caption{Models learned by applying SINDy with group sparsity regularization (sparsity parameter $\delta=1.5$) to each of the two ball drops.}
			\label{tab:learned-models}
		\end{table}

		The actual governing equations learned with the group sparsity method are provided in Table \ref{tab:learned-models}.
		Every equation has a constant acceleration term within a few meters per second squared of $-9.8$, but few are quite as close as one might expect.
		Thus even with a stable method of inferring governing equations, based on this data one would not necessarily conclude that all balls experience the same (mass-divided) force due to gravity.
		Note also that some of the balls mistakenly adopt \textit{positive} coefficients multiplying $v$.
		The balls for which this occurs tend to be those whose motion is well-approximated by constant acceleration.
		Because the size of the discrepancy between a constant acceleration model and these balls' measured trajectories is not much larger than the amount of error suspected to be present in the data, SINDy has a difficult time choosing an appropriate value for the $v$ terms.
		One would likely need higher resolution, higher accuracy measurement data in order to obtain reasonable approximations of the drag coefficients or $v^2$ terms.

		At 65 degrees Fahrenheit, the density of air $\rho$ at sea level is $1.211\text{kg}/\text{m}^3$ \cite{white2003fluid} and its dynamic viscosity $\mu$ is $1.82\times10^{-5}\text{kg}/(\text{m~s})$. The Reynolds number for a ball with diameter $D$ and velocity $v$ will then be
		\begin{equation}
			Re = 0.667 D v \times 10^5.
		\end{equation}
		Table \ref{tab:ball-stats} gives the maximum velocities of each ball over the two drops and the corresponding Reynolds numbers.
		Note that these are the \textit{maximum} Reynolds numbers, not the Reynolds numbers over the entire trajectories.
		With velocities under $30 \text{m}/\text{s}$ and diameters from $0.04\text{m}$ to $0.22\text{m}$ we should expect Reynolds numbers with magnitudes ranging from $10^4$ to $10^5$ over the course of the balls' trajectories (apart from the very beginnings of each drop).
		The \textit{average} trajectory consists of about 49 measurements, just over one of which corresponds to a Reynolds number that is $\mathcal{O}\left(10^3\right)$.
		About 13 of these measurements are associated with Reynolds numbers on the order of $10^4$ and roughly 33 with Reynolds numbers of magnitude $10^5$.
		Note that this means the majority of data points were collected when the balls were in the quadratic drag regime.
		Based on Figure \ref{fig:SphereDrag} we should expect balls with Reynolds numbers less than $10^5$ to have drag coefficients of magnitude about 0.5.
		Figure \ref{fig:SphereDrag} suggests that balls experiencing higher Reynolds numbers such as the volleyball and basketballs should have smaller drag coefficients varying between $0.05$ and $0.3$ depending on their smoothness.
		The predicted (linear) drag coefficients for the volleyball lie in this range while the basketballs' learned drag coefficients are erroneously positive.
		If the basketballs are treated as being smooth, their drag coefficients predicted by Figure \ref{fig:SphereDrag} may be too small for SINDy to identify given the noisy measurement data.
		A similar effect seems to occur for the golf ball. Though it experiences a lower Reynolds number, its dimples induce a turbulent flow over its surface, granting it a small drag coefficient at a lower Reynolds number.
		Overall, the linear drag coefficients predicted by the model are at least within a physically reasonable range, with some outliers having incorrect signs.

		\begin{table}[h]
			\centering
			\begin{tabular}{l|l|l|l|l|l}
			\textbf{Ball} & \textbf{Radius (m)} & \textbf{Mass (kg)} & \textbf{Density (kg/m)} & $v_{max}$\textbf{ (m/s)} & \textbf{Max Re}\\ \hline
				Golf Ball           & 0.021963 & 0.045359 & 1022.066427 & 26.63 & $1.75 \times 10^5$ \\ 
				Baseball            & 0.035412 & 0.141747 &  762.037525 & 26.61 & $2.83 \times 10^5$ \\ 
				Tennis Ball         & 0.033025 & 0.056699 &  375.813253 & 21.95 & $2.18 \times 10^5$ \\ 
				Volleyball          &  0.105*  & NA       &			 NA & 22.09 & $6.96 \times 10^5$ \\ 
				Blue Basketball     & 0.119366 & 0.510291 &   71.628378 & 24.80 & $8.88 \times 10^5$ \\ 
				Green Basketball    & 0.116581 & 0.453592 &   68.342914 & 25.06 & $8.77 \times 10^5$ \\ 
				Whiffle Ball 1      & 0.036287 & 0.028349 &  141.641937 & 16.91 & $1.84 \times 10^5$ \\ 
				Whiffle Ball 2      & 0.036287 & 0.028349 &  141.641937 & 16.35 & $1.78 \times 10^5$ \\ 
				Yellow Whiffle Ball & 0.046155 & 0.042524 &  103.250857 & 15.30 & $2.12 \times 10^5$ \\ 
				Orange Whiffle Ball & 0.046155 & 0.042524 &  103.250857 & 15.77 & $2.18 \times 10^5$ \\ 
			\end{tabular}
			\caption{Physical measurements, maximum velocities across the two drops, and maximum Reynolds numbers for the dropped balls. *We do not have measurement data for the volleyball, but obtained an estimate for its radius based on other volleyballs in order to approximate its maximum Reynolds number.}
			\label{tab:ball-stats}
		\end{table}

\begin{figure}
			\centering
			\includegraphics[width=\textwidth]{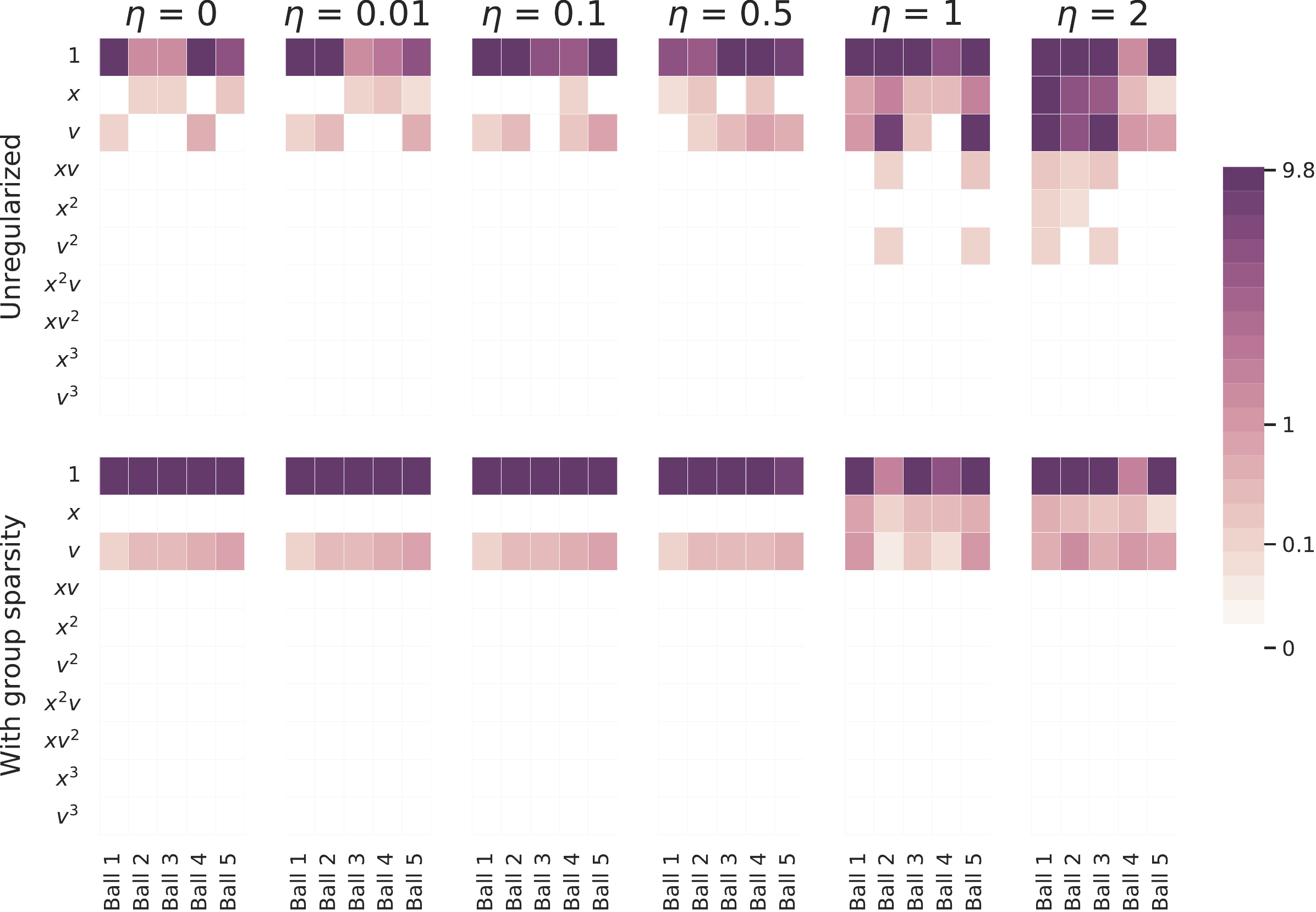}
			\caption{A comparison of coefficients of the models inferred from the simulated falling balls.
			The top row shows the coefficients learned with the standard SINDy algorithm and the bottom row the coefficients learned with the group sparsity method. $\eta$ indicates the amount of noise added to the simulated ball drops.
			The standard approach used a sparsity parameter of $0.05$ and the group sparsity method used a value of $1.5$.
			The balls were simulated using constant acceleration and the following respective coefficients multiplying $v$: $-0.1$, $-0.3$, $-0.3$, $-0.5$, $-0.7$.}
			\label{fig:synthetic_heatmaps}
		\end{figure}

		Next we turn to the simulated data set.
		We perform the same experiment as with the real world data: we apply both versions of SINDy to a series of simulated ball drops and then note the models that are inferred.
		Our findings are shown in Figure \ref{fig:synthetic_heatmaps}.
		We need not say much about the standard approach: it does a poor job of identifying coherent models for all levels of noise.
		The group sparsity regularization is much more robust to noise, identifying the correct terms and their magnitudes for noise levels up to half a meter (in standard deviation).
		For more significant amounts of noise, even this method is unable to decide between adopting $x$ or $v$ into its models.
		Perhaps surprisingly, if a $v^2$ term with coefficient $\sim0.1$ is added to the simulated model\footnote{It should be noted that, based on the balls' approximated velocities, the largest coefficient multiplying $v^2$ (i.e. $\tfrac{1}{2m}\rho A C_D$ from \eqref{Eq:fluiddrag}, where $m$ is the mass of a ball), is less than $0.08$ in magnitude, across all the trials.}, the learned coefficients look nearly identical.
		Although this additional term visibly alters the trajectory (before it is corrupted by noise), none of the learned equations capture it, even in the absence of noise.
		One reason for this is because the coefficient multiplying $v^2$ is too small to be retained during the sequential thresholding least squares procedure.
		If we decrease the sparsity parameter enough to accommodate it, the models also acquire spurious higher order terms.
		To infer the $v^2$ term using the approach outlined here, one would need to design and carry out additional experiments which better isolate this effect, perhaps by using a denser fluid or by dropping a ball with a larger diameter of relatively small mass, thereby increasing the constant multiplying $v^2C_D$ in \eqref{Eq:fluiddrag}.
		A much more realistic drag force based on \eqref{Eq:fluiddrag} can be used to simulate falling balls.
		Such a drag force will shift from being linear to quadratic in $v$ over the course of a ball's trajectory.
		In this scenario neither version of SINDy identifies a $v^2$ term, regardless of how much many measurements are collected, but both detect linear drag, exhibiting similar performance as is shown here.
		A more detailed discussion can be found in Appendix Section \ref{sec:realistic-simulations}.

	\subsection{Model error}\label{sec:model-error}
\begin{figure}
			\centering
			\includegraphics[width=\textwidth]{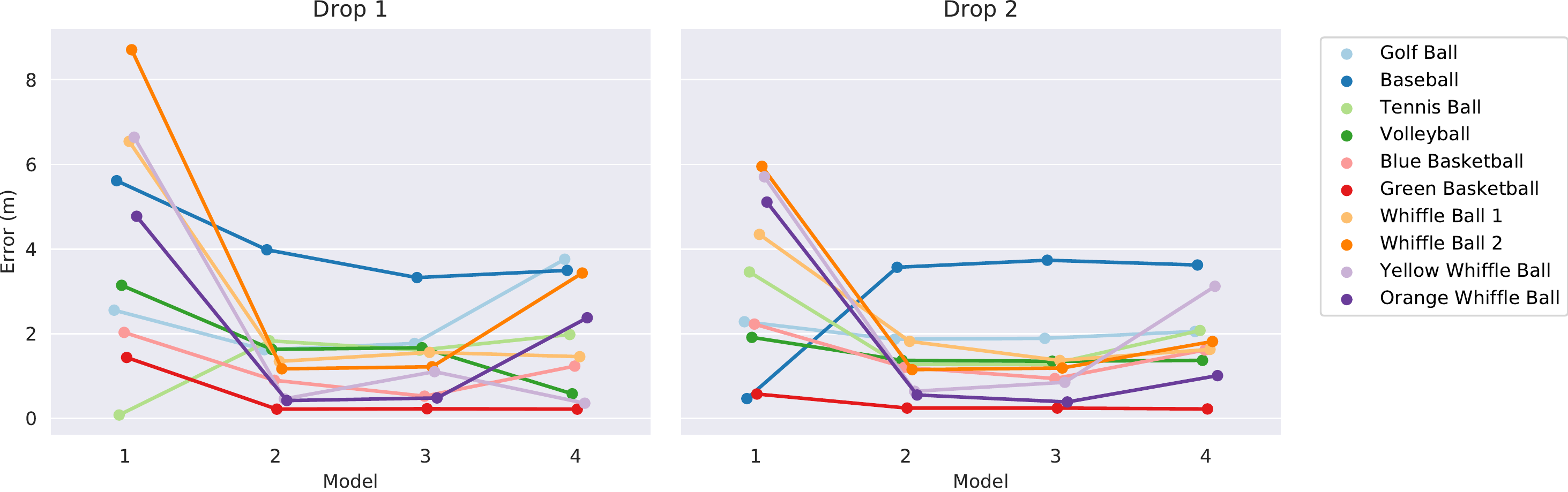}
			\caption{The error in landing time predictions for the four models. The results for the models trained on drops one and two are shown on the left and right, respectively. We have intentionally jittered the horizontal positions of the data points to facilitate easier comparison.}
			\label{fig:error_vs_model_real_jitter_darkgrid}
		\end{figure}

		We now turn to the problem of testing the predictive performance of models learned from the data.
		We benchmark four models of increasing complexity on the drop data. The model templates are as follows:

		\begin{enumerate}
			\item Constant acceleration: $\ddot x = \alpha$
			\item Constant acceleration with linear drag: $\ddot x = \alpha + \beta v$
			\item Constant acceleration with linear and quadratic drag: $\ddot x = \alpha + \beta v + \gamma v^2$
			\item Overfit model: Set a low sparsity threshold and allow SINDy to fit a more complicated model to the data
		\end{enumerate}
		The model parameters $\alpha, ~\beta,$ and $\gamma$ are learned using the SINDy algorithm using libraries consisting of just the terms required by the templates.
		The testing procedure consists of constructing a total of 80 models (4 templates $\times$ 10 balls $\times$ 2 drops) and then using them to predict a quantity of interest.
		First a template model is selected then it is trained using one ball's trajectory from one drop.
		Once trained, the model is given the initial conditions (initial height and velocity) from the same ball's other drop and tasked with predicting the ball's height after $2.8$ seconds have passed\footnote{This number corresponds to the shortest set of measurement data across all the trials. All models are evaluated at $2.8$ seconds to allow for meaningful comparison of error rates between models.}.
		Recall from Figure \ref{fig:landing_time_vs_density} that the same ball dropped twice from the same height by the same person on the same day can hit the ground at substantially different times.
		In the absence of any confounding factors, the time it takes a sphere to reach the ground after being released will vary significantly based on its initial velocity.
		Since there is sure to be some error in estimating the initial height and velocity of the balls, we should expect only modest accuracy in predicting their landing times.
		We summarize the outcome of this experiment in Figure \ref{fig:error_vs_model_real_jitter_darkgrid}.
		The error tends to decrease significantly between model one and model two, marking a large step in explaining the discrepancy between a constant acceleration model and observation.
		There does not appear to be a large difference between the predictive powers of models two and three as both seem to provide similar levels of accuracy.
		Occam's razor might be invoked here to motivate a preference for model two over model three since it is simpler and has the same accuracy.
		This provides further evidence that the level of noise and error in the data set is too large to allow one to accurately infer the dynamics due to $v^2$.
		Adding additional terms to the equations seems to weaken their generalizability somewhat, as indicated by the slight increase in errors for model four.

\begin{figure}
			\centering
			\includegraphics[width=\textwidth]{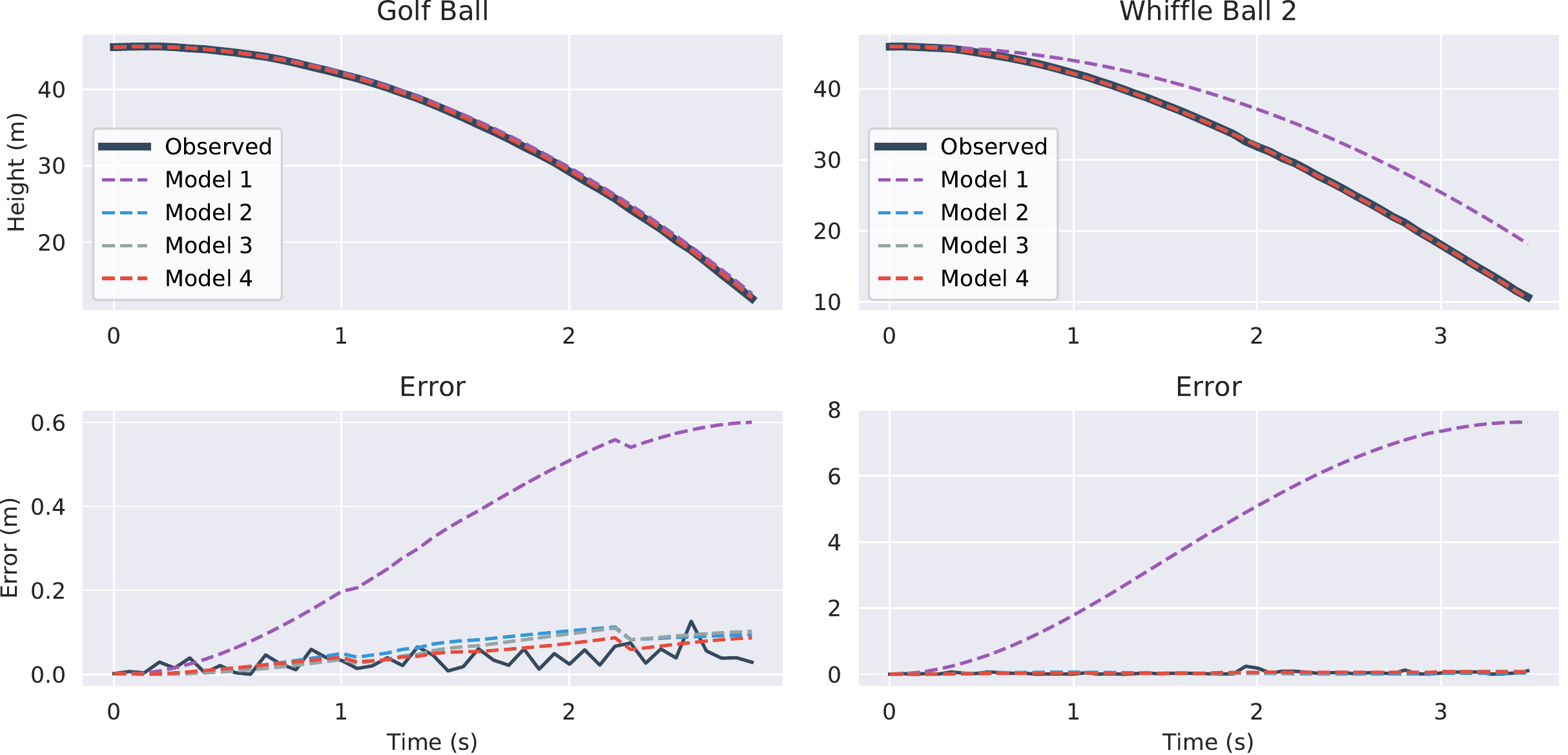}
			\caption{Predicted trajectories and error for the Golf Ball (top) and Whiffle Ball 2 (bottom). On the left we compare the predicted trajectories against the true path and on the right we show the absolute error for the predictions. The 'Observed' lines in the error plots show the difference between the original height measurements and the smoothed versions used for differentiation. They give an idea of the amount of intrinsic measurement noise. All models plotted were trained and evaluated on drop 2.}
			\label{fig:error_vs_time}
		\end{figure}

		Figure \ref{fig:error_vs_time} visualizes the forecasts of the learned equations for two of the balls along with their deviation from the true measurements.
		The models are first trained on data from drop 2, then they are given initial conditions from the same drop and made to predict the full trajectories.
		There are a few observations to be made.
		The constant acceleration models (model one) are clearly inadequate, especially for the whiffle ball.
		Their error is much higher than that of the other models indicating that they are underfitting the data, though constant acceleration appears to be a reasonable approximation for a falling golf ball. 
		Models two through four all seem to be imitating the trajectories to about the level of the measurement noise, which is about the most we could hope of them.
		It is difficult to say which model is best by looking at these plots alone.
		To break the tie we can observe what happens if we evaluate the models in ``unfamiliar'' circumstances and force them to extrapolate.

\begin{figure}
			\centering
			\includegraphics[width=\textwidth]{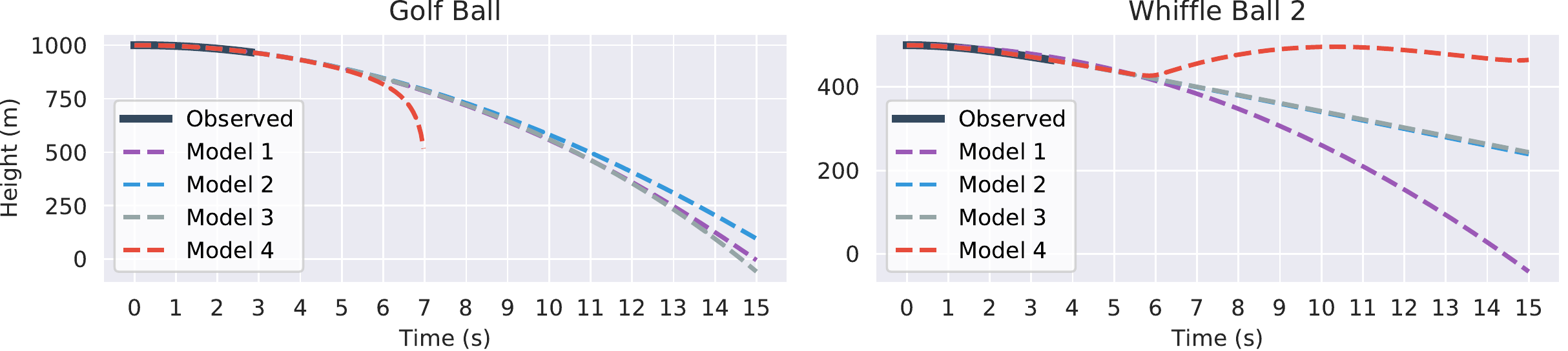}
			\caption{15 second forecasted trajectories for the Golf Ball (left) and Whiffle Ball 2 (right) based on the second drop. Part of the graph of Model 4 (red) is omitted in the Golf Ball plot because it diverged to $-\infty$.}
			\label{fig:long_time_predictions}
		\end{figure}

		Supplying the same initial conditions as before, with initial height shifted up to avoid negative heights, we task the models with predicting the trajectories out to 15 seconds.
		The results are shown in Figure \ref{fig:long_time_predictions}.
		All four models fit the observed data itself fairly well.
		However, six or seven seconds after the balls are released, a significant degree of separation has started to emerge between the trajectories.
		The divergence of the model four instances is the most abrupt and the most pronounced.
		The golf ball's model grows without bound after seven seconds.
		It is here that the danger of overfit, high-order models becomes obvious.
		In contrast, the other models are better behaved.
		For the golf ball models one through three agree relatively well, perhaps showing that it is easier to predict the path of a falling golf ball than a falling whiffle ball.
		That model two is so similar to the constant acceleration of model one also suggests that the golf ball experiences very little drag.
		The $v^2$ term for model three has a coefficient which is erroneously positive and essentially cancels out the speed dampening effects of the drag term, leading to an overly rapid predicted descent.
		Models two and three agree extremely well for the whiffle ball as the learned $v^2$ coefficient is very small in magnitude.

\section{Discussion and Conclusions}
	In this work, we have revisited the classic problem of modeling the motion of falling objects in the context of modern machine learning, sparse optimization, and model selection. 
	In particular, we develop data-driven models from experimental position measurements for several falling spheres of different size, mass, roughness, and porosity.  
	Based on this data, a hierarchy of models are selected via sparse regression in a library of candidate functions that may explain the observed acceleration behavior.  
	We find that models developed for individual ball-drop trajectories tend to overfit the data, with all models including a spurious height-dependent force and lower-density balls resulting in additional spurious terms.  
	Next, we impose the assumption that all balls must be governed by the same basic model terms, perhaps with different coefficients, by considering all trajectories simultaneously and selecting models via group sparsity.
	These models are all parsimonious, with only two dominant terms, and they tend to generalize without overfitting.

	Although we often view the motion of falling spheres as a solved problem, the observed data is quite rich, exhibiting a range of behaviors.  
	In fact, a constant gravitational acceleration is not immediately obvious, as the falling motion is strongly affected by complex unsteady fluid drag forces; the data alone would suggest that each ball has its own slightly different gravity constant.
	It is interesting to note that our group sparsity models include a drag force that is proportional to the velocity, as opposed to the \textit{textbook} model that includes the square of velocity that is predicted for a constant drag coefficient.
	However, in reality the drag coefficient decreases with velocity, as shown in Fig.~\ref{fig:SphereDrag}, which may contribute to the force being proportional to velocity. 
	Even when a higher fidelity drag model is used---a model containing rational terms missing from and poorly approximated by the polynomial library functions---to collect measurements uncorrupted by noise, SINDy struggles to identify coherent dynamics.
	In general SINDy may not exhibit optimal performance if not equipped with a library of functions in which dynamics can be represented sparsely.
	We emphasize that although the learned models tend to fit the data relatively well, it would be a mistake to assume that they would retain their accuracy for Reynolds numbers larger than those present in the training data.
	In particular we should expect the models to have trouble extrapolating beyond the drag crisis where the dynamics change considerably.
	This weakness is inherent in virtually all machine learning models; their performance is best when they are applied to data similar to what they have already seen and dubious when applied in novel contexts.
	That is to say they excel at interpolation, but are often poor extrapolators.

	Collecting a richer set of data should enable the development of refined models with more accurate drag physics\footnote{We note that in order to properly resolve these more complex drag dynamics with SINDy the candidate library would likely need to be enriched.}, and this is the subject of future work.
	In particular, it would be interesting to collect data for spheres falling from greater heights, so that they reach terminal velocity. 
	It would also be interesting to systematically vary the radius, mass, surface roughness, and porosity, for example to determine non-dimensional parameters.
	Finally, performing similar tests in other fluids, such as water, may also enable the discovery of added mass forces, which are quite small in air.
	Such a dataset would provide a challenging motivation for future machine learning techniques.

	We were able to draw upon previous fluid dynamics research to establish a ``ground truth'' model against which to compare the models proposed by SINDy.
	However, in less mature application areas one may not be fortunate enough to have a theory-backed set of reference equations, making it challenging to assess the quality of learned models.
	Many methods in numerical analysis come equipped with a priori or a posteriori error estimators or convergence results to give one an idea of the size of approximation errors.
	Similarly, in statistics goodness of fit estimators exist to help guide practitioners about what type of performance they should expect from various models.
	A comprehensive study into whether similar techniques could be adopted for application to SINDy would be an interesting topic for future research efforts.

	We believe that it is important to draw a parallel between great historical scientific breakthroughs, such as the discovery of a universal gravitational constant, and modern approaches in machine learning. 
	Although computational learning algorithms are becoming increasingly powerful, they face many of the same challenges that human scientists have faced for centuries.
	These challenges include trade offs between model fidelity and the quality and quantity of data, with inaccurate measurements degrading our ability to disambiguate various physical effects.
	With noisy data, one can only expect model identification techniques to uncover the dominant, leading-order effects, such as gravity and simple drag; for subtler effects, more accurate measurement data is required.  
	Modern learning architectures are often also prone to overfitting without careful cross-validation and regularization, and models that are both interpretable and generalizable come at a premium. 
	Typically the regularization encodes some basic human assumption, such as sparse regularization, which promotes parsimony in models. 
	More fundamentally, it is not always clear what should be measured, what terms should be modeled, and what parameters should be varied to isolate the effect one wishes to study.
	Historically, this type of scientific inquiry has been driven by human curiosity and intuition, which will be critical elements if machine intelligence is to advance scientific discovery.

\section*{Acknowledgments}
We would like to acknowledge funding support from the Defense Advanced Research Projects Agency (DARPA PA-18-01-FP-125) and the Air Force Office of Scientific Research (FA9550-18-1-0200 for SLB and FA9550-17-1-0329 for JNK).

\bibliographystyle{unsrt}
\bibliography{bibliography}

\newpage
\appendix
\appendixpage

\section{Sparse Identification of Nonlinear Dynamical systems revisited}
\label{sec:sindy-algo}
	
	In this section we provide some additional information concerning the Sparse Identification of Nonlinear Dynamical systems (SINDy) method.
	We first give a simplified implementation of the sequentially thresholded least-squares algorithm, implemented in Python, before showing examples of SINDy applied to two test problems: a nonlinear oscillator (Section \ref{sec:nonlinear-oscillator}) and a simulated falling body with different types of drag (Section \ref{sec:sindy-eq-of-motion}).

	Recall that to construct a set of governing equations, SINDy seeks to solve the following optimization problem
	\begin{equation}\label{eq:optimization-prob}
		\min_{\bXi}\frac12 \norm{\dot \X - \bPhi(\X)\bXi}_F^2 + \Omega(\bXi),
	\end{equation}
	where $\X$ is a matrix of measurements, $\dot \X$ is a matrix of derivatives of $\X$, $\bPhi(\X)$ is a library matrix whose columns consist of potential right-hand side functions evaluated on the measurement data, $\bXi$ is a coefficient matrix, and $\Omega(\cdot)$ is a regularization term encouraging sparsity.
	A one-dimensional version of the sequentially thresholded least-squares algorithm, which we use to solve \eqref{eq:optimization-prob} in this work\footnote{In actuality we use a custom implementation of sequentially thresholded least-squares for the main paper and most of the appendices and a recently developed package, PySINDy (\url{https://github.com/dynamicslab/pysindy}), for the examples in this section.}, can be implemented in Python as
	\begin{lstlisting}[language=iPython]
xi = least_squares(theta, x_dot)  # Initial guess

# delta is our sparsity parameter
for k in range(iterations):
    small_indices = abs(xi) < delta
    big_indices = ~small_indices
    
    xi[small_indices] = 0  # Threshold small coefficients
    xi[big_indices] = least_squares(x[:, big_indices], x_dot)
\end{lstlisting}
	Here we use \texttt{least\_squares} to denote a black-box least-squares solver. This implementation only solves for one column of the coefficients of $\bXi$, yielding the governing equation for only one measurement variable. In practice one runs this routine for each variable.

	\subsection{A brief example}\label{sec:nonlinear-oscillator}
\begin{figure}
			\centering
			\includegraphics[width=.5\textwidth]{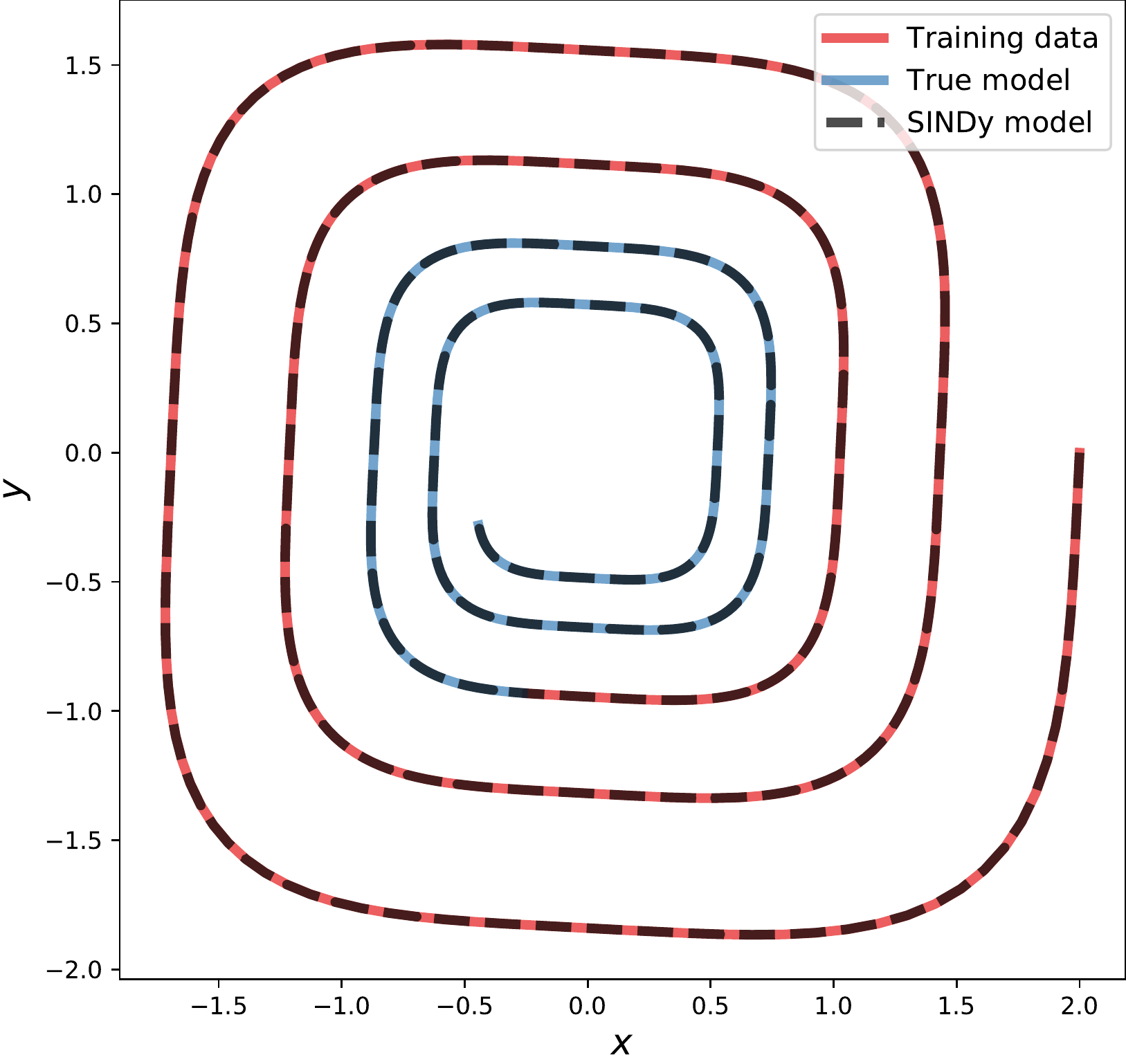}
			\caption{Dynamics of the nonlinear oscillator described by \eqref{eq:nonlinear-oscillator}. The true trajectory, computed using \eqref{eq:nonlinear-oscillator} is plotted as a solid line, with red denoting the training data fed to the SINDy model and blue denoting the portion of the trajectory unseen by SINDy. The dashed line shows the dynamics predicted by the model discovered by the SINDy model starting at initial condition $(2, 0)$.} 
			\label{fig:nonlinear_oscillator}
		\end{figure}

		Here we give an example of a dynamical system SINDy is easily able to identify: a first order nonlinear oscillator. The system is described by
		\begin{equation}\label{eq:nonlinear-oscillator}
			\begin{aligned}
				\dot x &= -\frac{1}{10}x^3 + 2y^3 \\
				\dot y &= -2 x^3 - \frac{1}{10}y^3.
			\end{aligned}
		\end{equation}
		To construct training data for a SINDy model we simulate a trajectory under these dynamics starting from $(2, 0)$ for $t\in [0, 5]$ with a time step of $0.01$. Using a threshold of $0.05$ and a library consisting of polynomials terms of degree up to five, SINDy recovers the following model
		\begin{align}
			\dot x &= -0.100 x^3 + 1.999 y^3 \\
			\dot y &= -1.999 x^3 - 0.100 y^3.
		\end{align}
		We plot the trajectories simulated from the actual model and the SINDy model for $t\in[0, 25]$ in Figure \ref{fig:nonlinear_oscillator}.
		Note the close agreement between the two trajectories.

	\subsection{Learning equations of motion}\label{sec:sindy-eq-of-motion}
		In this section we demonstrate that SINDy can readily learn simplified versions of the equations of motion, but struggles to identify dynamics containing terms not representable as linear combinations of the library terms.
		We simulate a ball of unit mass falling with constant acceleration and no drag
		\begin{equation}
			\dot v = -9.8, \qquad v(0) = 0
		\end{equation}
		and with constant acceleration and linear drag
		\begin{equation}
			\dot v = -9.8 - 0.5 v \qquad v(0) = 0.
		\end{equation}
		Each simulation consists of 50 height measurements taken every fifteenth of a second. We numerically differentiate the height data, then feed the velocity profiles to SINDy models with thresholds of $0.1$. SINDy learns the following governing equations:
		\begin{align}
			\dot v = -9.8000 &\qquad \text{(drag-free~simulation)}, \\
			\dot v = -9.786 - 0.499 v &\qquad \text{(linear~drag~simulation)}.
		\end{align}
		Multiple factors contribute to the accuracy of the learned models for these two test cases.
		The effects of the constant acceleration and drag on the ball trajectories are relatively large, the data lacks noise, and the appropriate terms are present in the trial libraries used by the SINDy models.

		On the other hand, when a higher fidelity drag model is used---one which contains terms missing from and poorly approximated by the library functions---SINDy struggles to identify coherent dynamics.
		Using the drag model given in \eqref{eq:force-ode} and \eqref{eq:CD-approx} (see Section \ref{sec:realistic-simulations}) to simulate a falling ball, SINDy learns the governing equation
		\begin{equation}
			\dot v = -6.345
		\end{equation}
		for a ``large'' threshold value ($0.1$).
		The constant acceleration term is shifted away from the true value to compensate for the drag.
		For a ``small'' threshold ($0.004$), SINDy learns the following model
		\begin{equation}
			\dot v = -9.810 - 0.005v + 0.17v^2.
		\end{equation}
		The constant acceleration is very close to the true value, but there is also a nonphysical positive quadratic term.
		Without including rational and other more complicated nonlinear functions in the library\footnote{Including rational functions in the library introduces additional complications to the SINDy algorithm \cite{Mangan2016ieee}.}, SINDy lacks the proper building blocks to perfectly reconstruct the behavior of the system.
		Poor performance can be a signal that some information is not being captured by the library, which is typically chosen based on one's underlying assumptions about the dynamics being studied.
		In this way SINDy can help reveal discrepancies between the assumed form of the governing equations and reality without necessarily exposing the precise nature of the discrepancy.
		If one finds that SINDy is producing unreliable models, a possible remedy is to enrich the library of candidate right-hand side functions.

\section{Numerical differentiation}
	\label{sec:numerical-differentiation-study}
	In this section we explore the error introduced by smoothing and numerical differentiation. More specifically in Section \ref{sec:diff-methods} we compare the performance of a few methods of numerical differentiation, in Section \ref{sec:smoothing} we examine the effects of smoothing on noisy data, and in Section \ref{sec:estimating-noise} we approximate the level of noise present in the actual ball drop data set and use the results of the previous sections to derive estimates for the error in the numerical derivatives used in the paper.

\begin{figure}
		\centering
		\includegraphics[width=.85\textwidth]{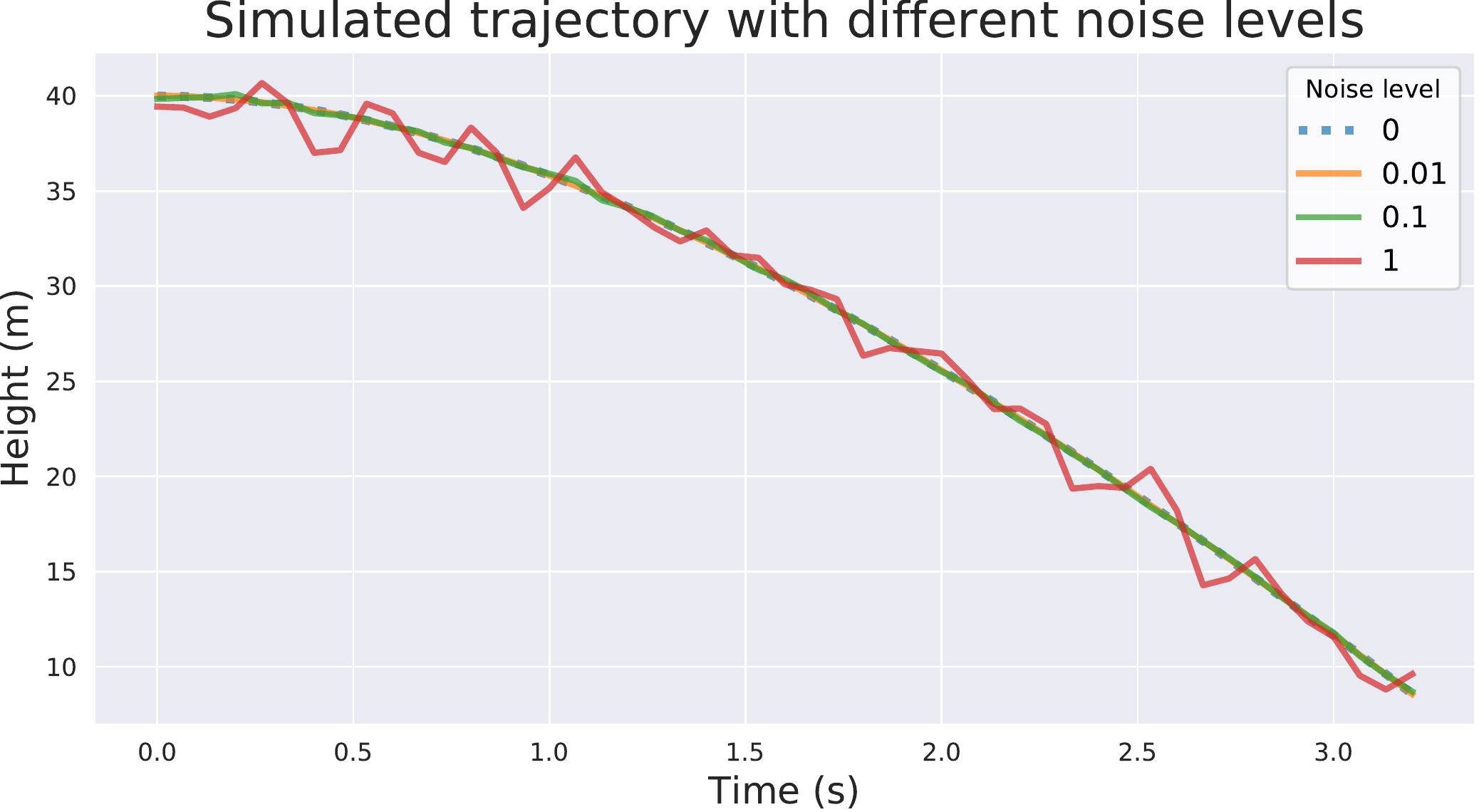}
		\caption{The simulated trajectory used for our numerical differentiation and smoothing experiments with varying amounts of noise added.} 
		\label{fig:noisy_trajectory}
	\end{figure}

	Unless otherwise noted, we worked with a single synthetic trajectory consisting of height measurements generated from an idealized falling object obeying
	\begin{equation}\label{eq:height-model}
		\dot v = -9.8 -0.5 v, \qquad v(0) = 0,\qquad x(0) = 40.
	\end{equation}
	This particular model was chosen because it is qualitatively similar to the actual trajectories. The measurements are taken at a rate of 15 per second to further imitate the experimental setup. We then add various amounts of Gaussian noise to the measurements. In the plots that follow ``noise level'' refers to the standard deviation of the noise added. Figure \ref{fig:noisy_trajectory} shows the trajectory with various amounts of noise. Note that even a noise level of $0.1$ is almost indistinguishable from the true trajectory.

	\subsection{Differentiation method comparison}\label{sec:diff-methods}
		We evaluate four numerical differentiation variants: two (first order) forward difference methods and two (second order) centered difference methods. For one method of each order we apply Savitzgy-Golay smoothing before performing computing the derivative. For the remaining two methods (one first order and one second order) we do not smooth the data before taking the derivative. We use a window size of 35 when performing smoothing. We compute both the first derivative (velocity) and second derivative (acceleration) of the simulated trajectory since the associated differential equation is second order.

		\begin{figure}
			\centering
			\includegraphics[width=\textwidth]{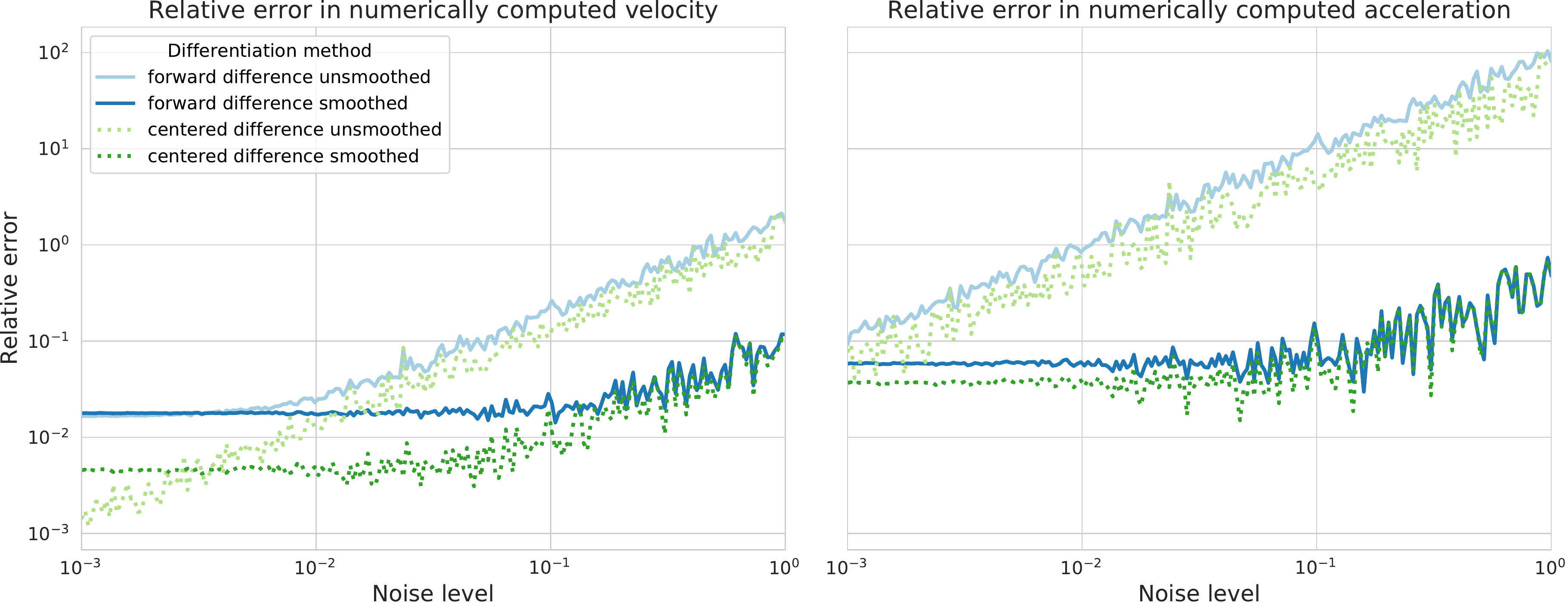}
			\caption{Left: Relative error in the {\it first} derivative of the trajectory computed using four differentiation methods with varying amounts of noise. Right: Relative error in the {\it second} derivative of the trajectory computed using four differentiation methods with varying amounts of noise.} 
			\label{fig:differentiation_method_comparison}
		\end{figure}

		Figure \ref{fig:differentiation_method_comparison} summarizes our results. There are a few observations to be made:
		\begin{itemize}
			\item The smoothed versions of the methods exhibit much better performance than the unsmoothed variants as the noise level increases.
			\item Once enough noise is introduced, all the methods considered see their accuracy degraded roughly linearly with the noise level.
			\item The error levels are higher for the approximate acceleration than for the velocity. This makes sense since some error is introduced in computing the velocity and the velocity is needed to compute the acceleration.
			\item At a low enough noise level there tends to be little difference between the smoothed and unsmoothed versions of each method. The unsmoothed centered difference method outperforms its smoothed counterpart in computing the velocity of relatively clean data.
		\end{itemize}
		
		A conclusion we can draw from this analysis is that the smoothed centered difference method provides the best performance over most levels of noise for both the first and second derivatives.

	\subsection{Smoothing}\label{sec:smoothing}
		In the previous experiment we used a fixed window length without justifying our choice. In this section we fix the differentiation method used --- centered difference with smoothing --- and vary the window length. A larger window means that more points are considered when performing smoothing. The window length roughly translates to smoothness; the larger the window the smoother the result.

		\begin{figure}
			\centering
			\includegraphics[width=\textwidth]{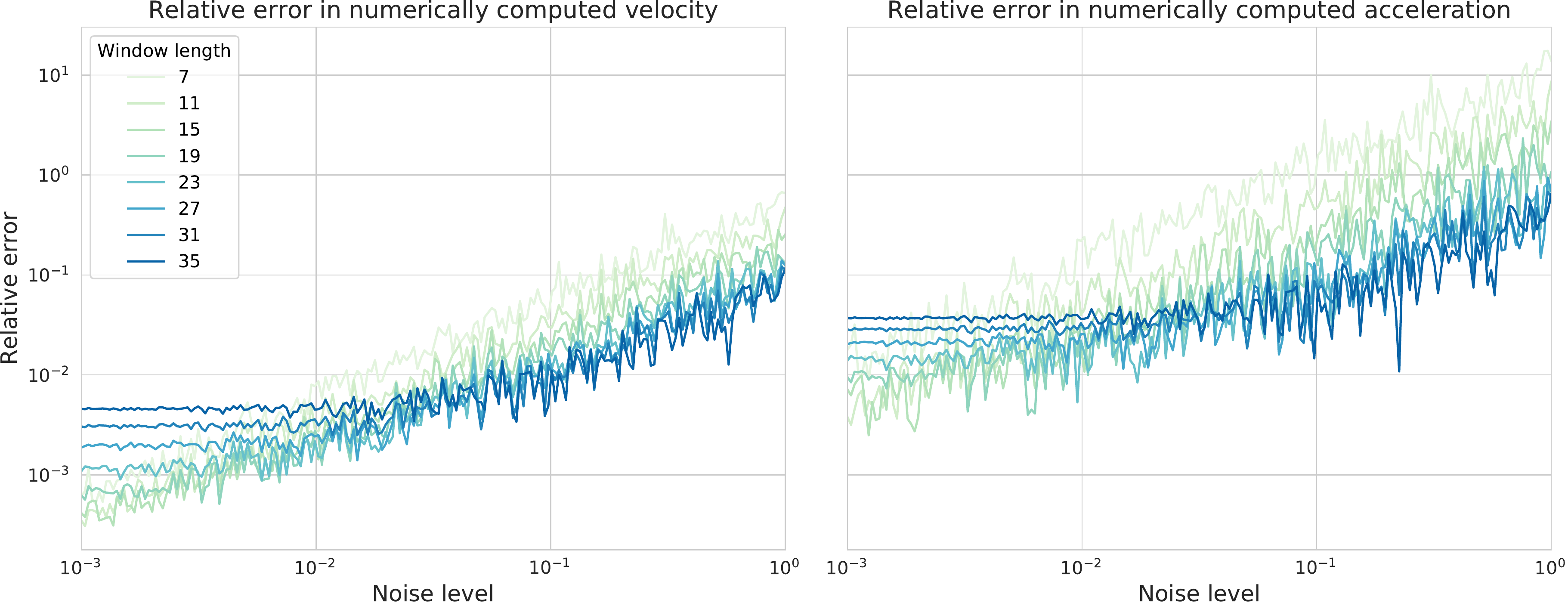}
			\caption{The effects of the size of the smoothing window on derivative approximation error. Left: Relative error in the {\it first} derivative of the trajectory computed using a smoothed centered difference method with different smoothing window sizes. Right: Relative error in the {\it second} derivative of the trajectory computed using a smoothed centered difference method with different smoothing window sizes. } 
			\label{fig:differentiation_smoothing_window_comparison}
		\end{figure}

		Figure \ref{fig:differentiation_smoothing_window_comparison} plots how the error in numerically computed derivatives is affected by the size of the smoothing window used as a function of noise. For small noise levels, larger smoothing windows hurt the method; overly aggressive smoothing throws out some useful information. As the noise levels increase the opposite is true; larger amounts of smoothing are needed to keep the excessive noise at bay. Which window length we should actually use will depend on the noise level we suspect is present in the real-world data set.

	\subsection{Estimating noise in measurement data}\label{sec:estimating-noise}
	In order to infer the amount of noise in the measured ball trajectories it will prove useful to know roughly how much the act of smoothing a trajectory changes the underlying data. To this end we perform a similar experiment as in the previous section, but with the height data itself. That is to say we apply the same smoothing operation used before to the height data and measure the relative difference between the smoothed and original data. For completeness, we carry out this experiment for multiple window lengths.

	\begin{figure}
		\centering
		\includegraphics[width=.85\textwidth]{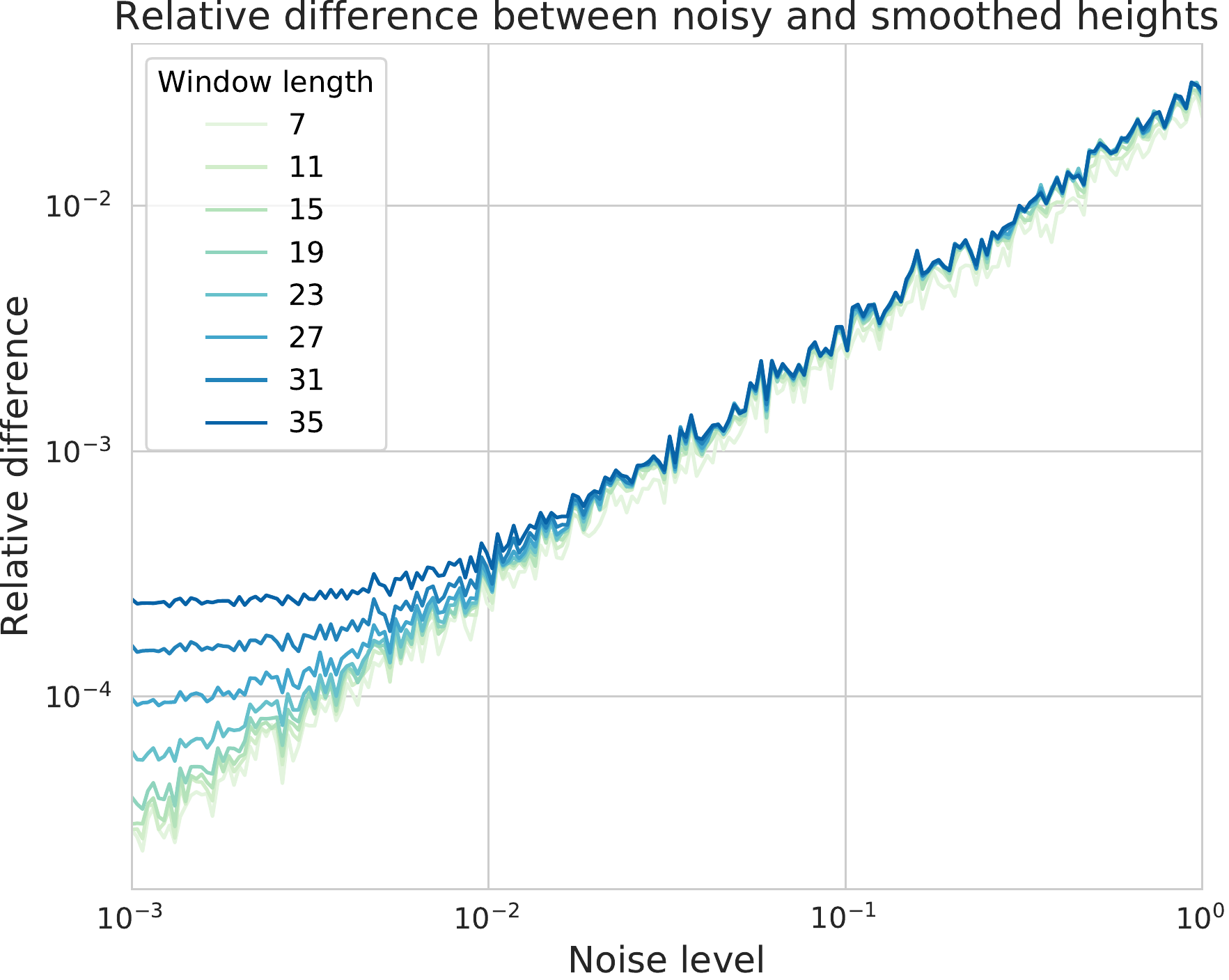}
		\caption{The relative difference between noisy trajectories and their smoothed versions for different length smoothing windows.} 
		\label{fig:smoothing_error}
	\end{figure}

	Our results are shown in Figure \ref{fig:smoothing_error}. As one would expect, smaller windows produce smoothed trajectories that are closer to their unsmoothed counterparts, but only slightly so. Large smoothing windows have the most pronounced effects when the noise levels are very small and smoothing is unnecessary. For higher noise levels, changes in window size affect the relative difference very little.

	Based on these results we elect to use the largest window size tested in our experiments in the main paper. SINDy depends heavily on accurate numerical derivatives. For large amounts of noise, a larger window size is necessary for numerical differentiation to work well. We are only penalized for using a large window (in the sense that we greatly modify the original data when we perform smoothing) if the underlying noise level is below about $10^{-2}$.

\begin{figure}
		\centering
		\includegraphics[width=\textwidth]{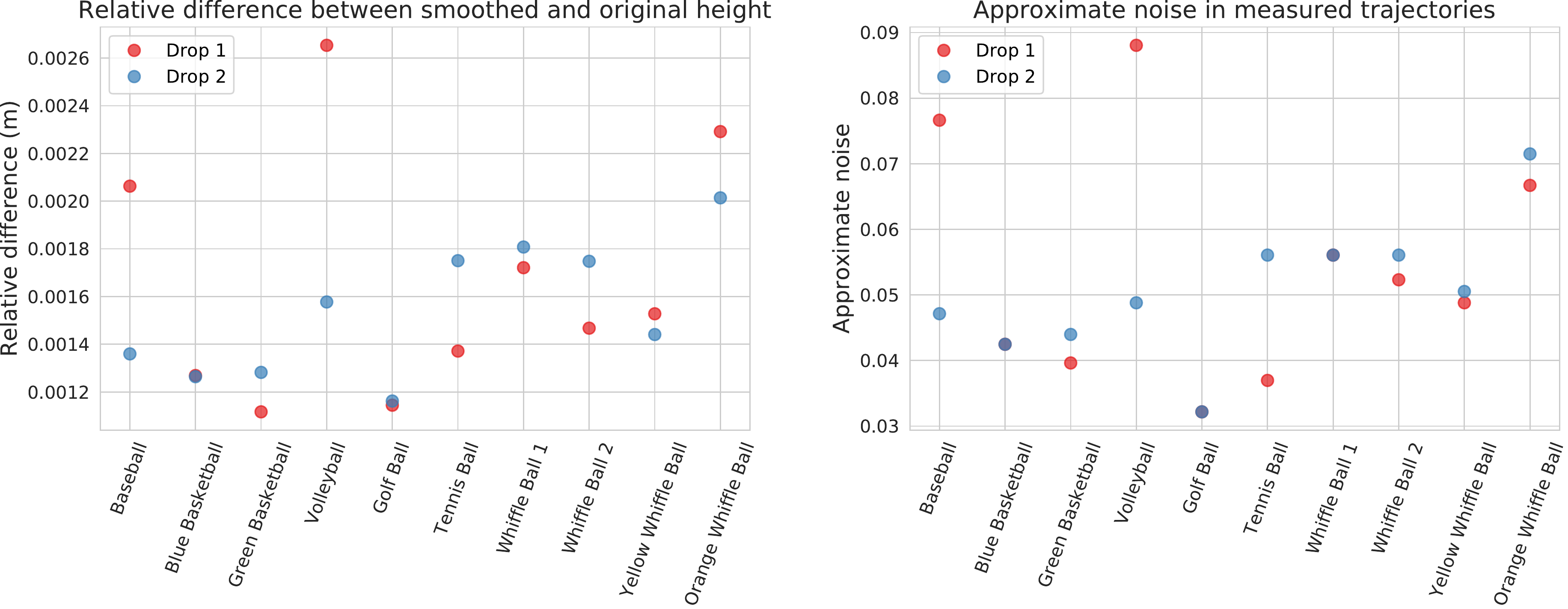}
		\caption{Left: Relative difference between the smoothed and unsmoothed falling ball trajectories for both drops (window length = 35). Right: Approximate noise levels present in the ball drop measurements.} 
		\label{fig:noise_estimation}
	\end{figure}

	Next we turn to the task of actually estimating the noise present in the drop data. To accomplish this we apply smoothing with a window length of 35 to each falling ball trajectory, then measure the relative difference between the smoothed and unsmoothed versions. Finally we compare this relative difference with Figure \ref{fig:smoothing_error} to obtain an approximation to the noise level. Figure \ref{fig:noise_estimation} visualizes the relative differences along with the inferred noise levels for each ball drop. The estimated noise levels are all between $0.035$ and $0.065$. Comparing these results with Figure \ref{fig:differentiation_method_comparison}, we can deduce that the numerically computed velocity and acceleration vectors have relative errors of order $10^{-3}$ and $10^{-2}$, respectively. It should be noted that we use the $\ell^2$ norm when computing relative error. If the $\ell^\infty$ norm is used instead, the relative errors increase marginally.

\section{Effect of varying sparsity parameter}
\label{sec:sparsity-paramater-study}
	In this section we provide a representative example of the that are produced when the sparsity threshold parameter is varied.
	We consider both the regularized and unregularized SINDy variants. The effect of varying the sparsity parameter is similar In both cases. For large values of the parameter (reflecting a strong preference toward a very sparse solution), no terms are deemed ``important'' enough to be retained and the trivial model is returned. As the threshold is continuously decreased, a small number of terms will be selected for a range of threshold values. Eventually, when the threshold becomes small enough, suddenly there will be a noticeable jump in the number of terms in the models returned by SINDy. This is typically when one can assume that the sparsity parameter has been made too small. We demonstrate this pattern in Tables \ref{tab:unreg-eqns} and \ref{tab:reg-eqns}, which give the learned equations for unregularized and unregularized SINDy, respectively, for a variety of sparsity thresholds. The parameter values were chosen to be close to values at which the number of terms in the resulting models changed. For the experiments carried out in the paper we chose thresholds which were slightly larger than the values at which the jumps in numbers of model terms occurred.

	\begin{table}[h]
		\centering
		\begin{tabular}{l|l}
			{\bf Threshold} & {\bf Equation}\\ \hline
			10     & $v'  =  0 $                                                                                    \\
			2      & $v'  =  -7.6344 $                                                                              \\
			0.1    & $v'  =  -14.865 +  0.1084 x - 0.2914 v $                                                       \\
			0.005  & $v'  =  -6.1068 -  0.0717 x + 0.0880 v -  0.0059 xv $                                          \\
			0.0045 & $v'  =  -2.9116 -  0.1388 x + 0.0861 v -  0.0061 xv - 0.0048 v^2 $                             \\
			0.0035 & $v'  =  14.7998 -  0.6964 x + 0.7036 v -  0.0182 xv + 0.0039 x^2 -  0.0065 v^2 $               \\
			0.002  & $v'  =  45.4998 -  1.4749 x + 2.4829 v -  0.0559 xv + 0.0067 x^2 -  0.0393 v^2 -  0.0021 v^3 $ \\
		\end{tabular}
		\caption{Models learned by unregularized SINDy for different threshold parameters (tennis ball, drop one).}
		\label{tab:unreg-eqns}
	\end{table}

	\begin{table}[h]
		\centering
		\begin{tabular}{l|l}
			{\bf Threshold} & {\bf Equation}\\ \hline
			70	   & $v'  =  0 $ \\
			65	   & $v'  =  -6.9 $ \\
			2	   & $v'  =  -8.3 -  0.1 v $ \\
			0.2	   & $v'  =  -15.6 + 0.1 x -  0.3 v $ \\
			0.14   & $v'  =  -2.0 -  0.1 x +  0.4 v - 0.01 xv $ \\
			0.1	   & $v'  =   1.5 -  0.2 x +  0.4 v - 0.01 xv +  0.001 v^2 $ \\
			0.05   & $v'  =  -13.1 + 0.2 x -  0.6 v + 0.008 xv -  0.003 x^2 -  0.009 v^2 $ \\
			0.02   & $v'  =   21.7 - 0.7 x +  1.7 v - 0.04 xv +  0.002 x^2 -  0.05 v^2 -  0.003 v^3 $ \\
			0.01   & $v'  =  -35.2 + 1.1 x -  4.7 v + 0.13 xv -  0.01 x^2 -  0.2 v^2 -  0.0009 x^2v + 0.003 xv^2 -  0.001 v^3 $ \\
			0.005  & {$\!\begin{aligned}
						v'  &=  9.9 -  0.8 x - 0.8 v -  0.04 xv + 0.02 x^2 -  0.005 v^2 + 0.0004 x^2v - 0.0004 xv^2 \\ &- 0.0002 x^3 -  0.0003 v^3
					 \end{aligned}$} \\
		\end{tabular}
		\caption{Models learned by regularized SINDy for different threshold parameters (all balls, drop one).}
		\label{tab:reg-eqns}
	\end{table}

	In cases where one wishes to perform automatic parameter tuning, cross-validation should allow for one to choose an appropriate sparsity parameter value. Models that are overly sparse (those which have too large a sparsity parameter) will be too simple to accurately predict unseen data and models that are not sparse enough (those which have too small a sparsity parameter) will overfit the training data and will generalize poorly. Poor performance on the holdout/validation/test set should catch both overfit and underfit models.

	One troublesome case that is possible with SINDy is when the model jumps directly from underfitting to overfitting as the sparsity parameter is varied. This could occur for a number of reasons, but the primary suspects are typically:
	\begin{itemize}
		\item The library is not rich enough to properly capture the dynamics (i.e. one or more of the terms in the ``true'' underlying dynamical system are not present in the library being used by SINDy).
		\item The library is too rich. If too many functions are included in the library then the system solved by SINDy can become ill-conditioned, leading to unpredictable results.
		\item The data are not described by a dynamical system. If this is the case then SINDy is not an appropriate tool.
		\item Data are too noisy. Recovering a sparse solution from extremely noisy data may not be possible.
	\end{itemize}

\section{Realistic falling ball simulations}
	\label{sec:realistic-simulations}

\begin{figure}
		\centering
		\includegraphics[width=\textwidth]{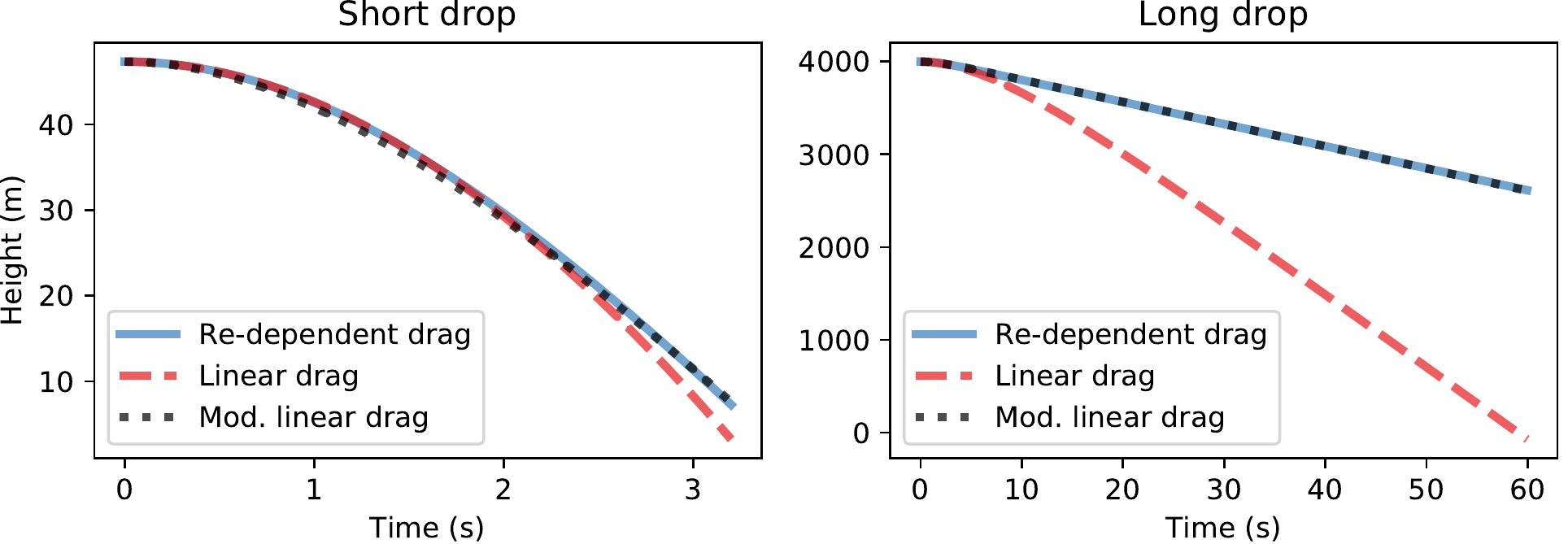}
		\caption{A comparison of the simulated trajectories of a tennis ball using a Reynolds number-dependent drag force from \eqref{eq:force-ode} and \eqref{eq:CD-approx} (solid) and two different constant linear drags (dashed and dotted). On the left we show a short drop similar to the physical experiments and on the right we have simulated a longer drop lasting a full 60 seconds. For the Reynolds number-dependent drag model we used the mass and diameter of the actual tennis ball. For the first linear drag model we used constant gravitational acceleration and a drag coefficient of $-0.125$ (the average of the two drag coefficients selected by SINDy in the real-world experiments). The modified linear drag model involved constant acceleration of $-12.7~\text{m}/\text{s}^2$ and a drag coefficient of $-0.53$. No noise was added.}
		\label{fig:drag_comparison}
	\end{figure}

	In the main work we simulate falling balls with \textit{constant} drag proportional to the balls' velocities.
	However, in reality, the drag varies nonlinearly with Reynolds number, which it itself a function of velocity.
	In this section we discuss what the two SINDy models are able to learn when a more complicated, but physically accurate model is used to construct the synthetic ball drops.
	Specifically, to simulate falling spheres, we numerically solve the following initial value problem for 49 time steps of length $1/15$ seconds (to mimic the real-world experiments)
	\begin{equation}\label{eq:force-ode}
		m\dot v = mg + \frac12\rho v^2 A C_D,\qquad v(0) = 0
	\end{equation}
	where $m$ is the mass of the ball, $\rho$ is the density of air, $A=\pi r^2$ is the cross-sectional area of the sphere, $r$ is the radius of the sphere, and $C_D$ is the Reynolds number dependent drag coefficient. 
	We use $\rho=1.211~\text{kg}/\text{m}^3$ (the density of air at sea level with a temperature of 65 degrees Fahrenheit).
	For $C_D$ we use the following approximation which is based on experimental measurements, recommended by Brown and Lawler \cite{brown2003sphere}:
	\begin{equation}\label{eq:CD-approx}
	 	\frac{24}{Re}\left(1 + 0.150Re^{0.681}\right) + \frac{0.407}{1 + \frac{8710}{Re}}.
	\end{equation}
	This approximation is valid for $Re < 2\times 10^5$, just before the so-called ``drag crisis'' when $C_D$ drops suddenly.
	We do not attempt to reproduce the behavior of the drag coefficient during and after the drag crisis, only before it. 
	This model also assumes that the spheres are smooth.
	In Figure \ref{fig:drag_comparison} we compare the simulated trajectories for a tennis ball using both the drag force from \eqref{eq:force-ode} and \eqref{eq:CD-approx} and the linear drag model presented in the main paper. Note that the trajectories initially agree, but as the ball reaches higher velocities and Reynolds numbers, the more complicated $Re$-dependent model predicts a larger drag force.
	The difference between these two models becomes clear when the simulations are run for longer amounts of time (900 time steps). 
	The ball effected by linear drag reaches a much faster terminal velocity compared with the other ball.
	Figure \ref{fig:drag_comparison} also shows how a linear drag model with a larger drag coefficient (and larger constant acceleration) can mimic the $Re$-dependent model.

\begin{figure}
		\centering
		\includegraphics[width=\textwidth]{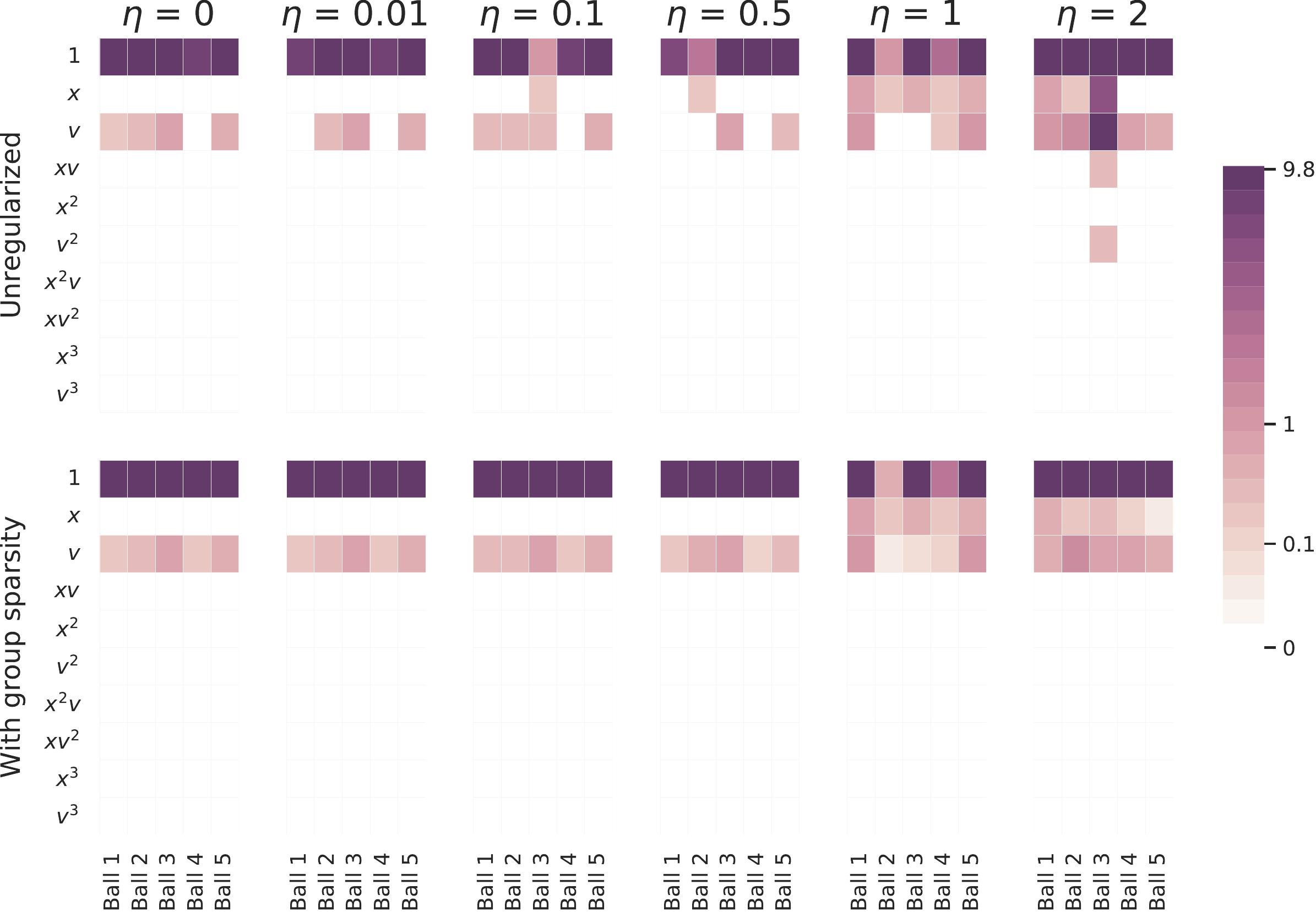}
		\caption{A comparison of coefficients of the models inferred from the simulated falling balls. The top row shows the coefficients learned with the standard SINDy algorithm and the bottom row the coefficients learned with the group sparsity method. $\eta$ indicates the amount of noise added to the simulated ball drops.
		The standard approach used a sparsity parameter of $0.05$ and the group sparsity method used a value of $0.3$.
		The balls' trajectories were simulated using equation \eqref{eq:force-ode}.} 
		\label{fig:synthetic_heatmaps_realistic_drag}
	\end{figure}

	As before we simulate five idealized balls falling for 49 time steps of duration $1/15$ seconds, each with a different mass and radius.
	The masses and radii were selected to match a subset of the balls in the real-world data set. Table \ref{tab:ball-properties} gives the characteristics of each simulated ball.
	Varying amounts of noise are then added to the artificial measurement data.
	Finally, we apply the unregularized and group variants of SINDy.
	The coefficients learned by the two methods are shown in Figure \ref{fig:synthetic_heatmaps_realistic_drag}.

	\begin{table}[h]
		\centering
		{
			\begin{tabular}{l|l|l|l}
				\textbf{Simulated ball} & \textbf{Real ball} & \textbf{Radius (m)} & \textbf{Mass (kg)}	\\ \hline
				Ball 1 & Golf Ball 		& 0.022 & 0.0454 \\
				Ball 2 & Tennis Ball 	& 0.033 & 0.0567 \\
				Ball 3 & Whiffle Ball 1 & 0.036 & 0.0283 \\
				Ball 4 & Baseball 		& 0.035 & 0.1417 \\
				Ball 5 & Blue Basketball& 0.119 & 0.5103 \\
			\end{tabular}
		}
		\caption{Properties of the simulated balls and the real balls after which they were modeled.}
		\label{tab:ball-properties}
	\end{table}

	The unregularized SINDy models exhibit better performance here than in the linear drag case in the sense that they tend not to pick up extraneous terms such as $x$ until relatively high noise levels are present.
	Notably, many of the models include constant acceleration and linear drag terms.
	The group sparsity methods perform similarly as before.
	For low noise levels it detects constant acceleration and linear, but not quadratic drag.
	As additional noise is introduced, the models erroneously adopt a term proportional to ball height.
	It should be noted that, in these simulations, the factor multiplying $v^2$ in \ref{eq:force-ode}, namely $\tfrac{1}{2m}\rho A C_D$, does not exceed $0.08$, except very early in the balls' trajectories when $v$ is small and $v^2$ even smaller.
	The consequence of this observation is that even if this factor were constant with respect to velocity, SINDy and other model discovery methods would have a difficult time accurately detecting it because it is so small relative to the other effects present in the experiment.
	It should be noted that even if the number of measurements is expanded by increasing the duration of the simulations, SINDy tends to adjust the constant acceleration and linear drag coefficients to match the data rather than adopting a quadratic drag term.
	Figure \ref{fig:drag_comparison} demonstrates just how closely linear drag can mimic quadratic drag as a ball approaches terminal velocity.
	If the amount of data is increased by instead collecting more measurements over a shorter time span\footnote{We experimented with increasing the sampling rate to 60 measurements per second over 3.33 seconds.} we saw no improvements in the ability of the model to detect a quadratic drag term.
	Similarly, SINDy accounted for increases in the density of the fluid through which the balls fall by adjusting the gravitational constant and linear drag term.

\end{document}